\documentclass{article}

 \PassOptionsToPackage{numbers, compress}{natbib}
\usepackage{graphicx}
\usepackage{caption}
\usepackage{subcaption}
\graphicspath{{./photographs/}{./fig/}}
\usepackage{multirow}
\usepackage{amsmath}
\usepackage{setspace}
\usepackage{algorithm}
\usepackage[noend]{algorithmic}

\usepackage[inline,shortlabels]{enumitem}
\usepackage[final]{nips_2017}

\usepackage[utf8]{inputenc} 
\usepackage[T1]{fontenc}    
\usepackage{hyperref}       
\usepackage{url}            
\usepackage{booktabs}       
\usepackage{amsfonts}       
\usepackage{nicefrac}       
\usepackage{microtype}      

\title{Active Bias: Training More Accurate Neural Networks by Emphasizing High Variance Samples}

\author{
Haw-Shiuan Chang,  Erik Learned-Miller,  Andrew McCallum \\
University of Massachusetts, Amherst \\
 140 Governors Dr., Amherst, MA 01003 \\
  \texttt{\{hschang,elm,mccallum\}@cs.umass.edu} \\
}

\begin{document}

\maketitle

\begin{abstract}
{\em Self-paced learning} and {\em hard example mining} re-weight training instances to improve learning accuracy. This paper presents two improved alternatives based on lightweight estimates of sample uncertainty in stochastic gradient descent (SGD): the variance in predicted probability of the correct class across iterations of mini-batch SGD, and the proximity of the correct class probability to the decision threshold. Extensive experimental results on six datasets show that our methods reliably improve accuracy in various network architectures, including additional gains on top of other popular training techniques, such as residual learning, momentum, ADAM, batch normalization, dropout, and distillation.
\end{abstract}


\section{Introduction}
\label{sec:intro}
Learning easier material before harder material is often beneficial to human learning. Inspired by this observation, {\em curriculum learning}~\cite{bengio2009curriculum} has shown that learning from easier instances first can also improve neural network training. When it is not known {\em a priori} which samples are easy, examples with lower loss on the current model can be inferred to be easier and can be used in early training. This strategy has been referred to as {\em self-paced learning}~\cite{kumar2010self}. By decreasing the weight of difficult examples in the loss function, the model may become more robust to outliers~\cite{meng2015objective}, and this method has proven useful in several applications, especially with noisy labels~\cite{pi2016self}.

Nevertheless, selecting easier examples for training often slows down the training process because easier samples usually contribute smaller gradients, and the current model has already learned how to make correct predictions on these samples. On the other hand, and somewhat ironically, the opposite strategy (i.e., sampling harder instances more often) has been shown to accelerate (mini-batch) stochastic gradient descent (SGD) in some cases, where the difficulty of an example can be defined by its loss~\cite{hinton2007recognize,loshchilov2015online,shrivastava2016training} or be proportional to the magnitude of its gradient~\cite{ZhaoZ14,alain2015variance,gao2015active,gopal2016adaptive}. This strategy is sometimes referred to as {\em hard example mining}~\cite{shrivastava2016training}.


In the literature, we can see that these two opposing strategies work well in different situations. Preferring easier examples may be effective when either machines or humans try to solve a challenging task containing more label noise or outliers. On the other hand, focusing on harder samples may accelerate and stabilize SGD in cleaner data by minimizing the variance of gradients~\cite{alain2015variance,gao2015active}. However, we often do not know how noisy our training dataset is. Motivated by this practical need, this paper explores new methods of re-weighting training examples that are effective in both scenarios.


\begin{figure}[t]
\begin{center}
\centerline{\includegraphics[width=0.6\columnwidth]{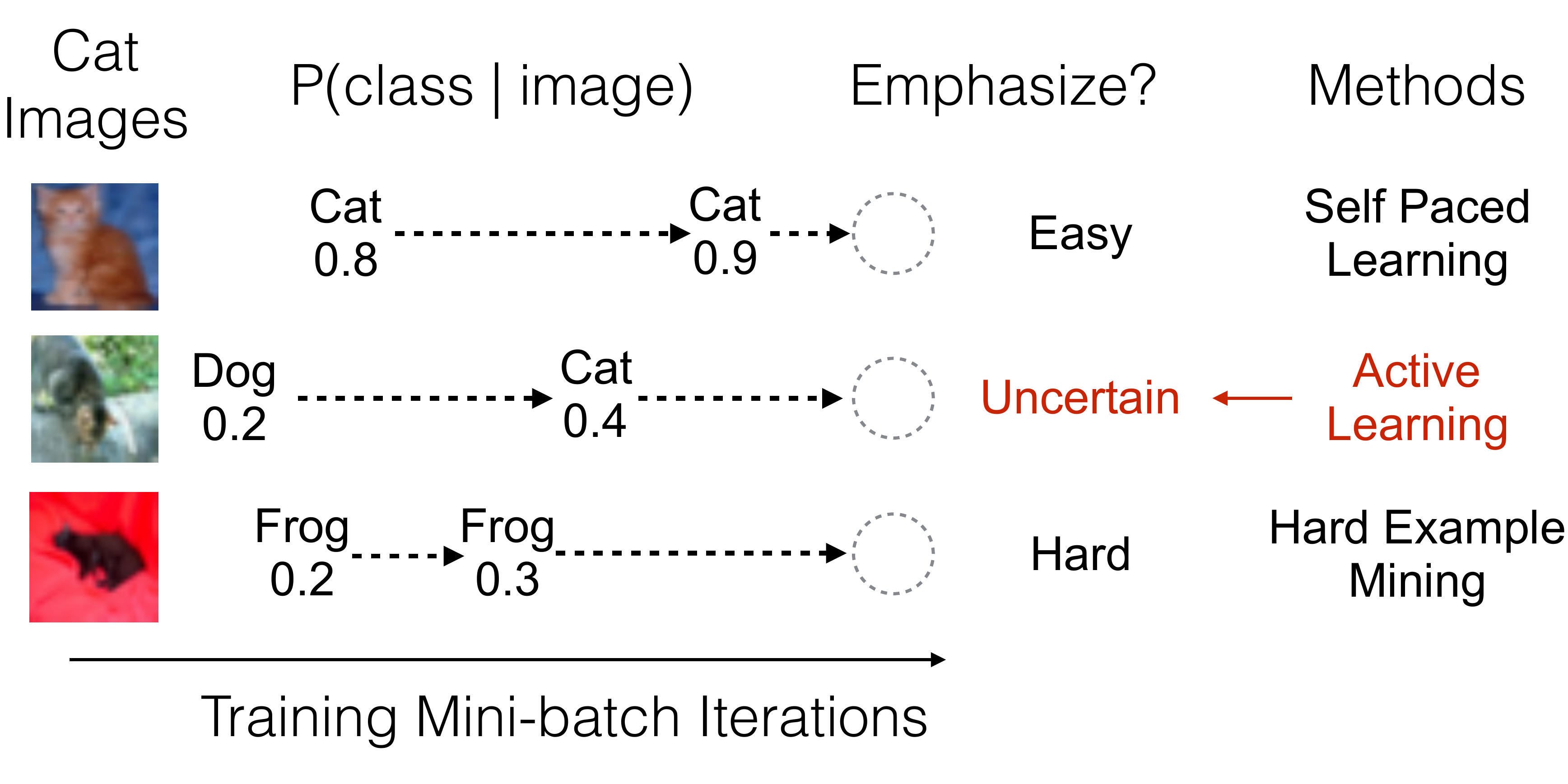}}
\vspace{-1mm}
\caption{The proposed methods emphasize uncertain samples based on previous prediction history.}
\label{fig:first_page}
\end{center}
\vspace{-3mm}
\end{figure} 

Intuitively, if a model has already predicted some examples correctly with high confidence, those samples may be too easy to contain useful information for improving that model further. Similarly, if some examples are always predicted incorrectly over many iterations of training, these examples may just be too difficult/noisy and may degrade the model. This suggests that we should somehow prefer uncertain samples that are predicted incorrectly sometimes during training and correctly at other times, as illustrated in~Figure~\ref{fig:first_page}. This preference is consistent with common variance reduction strategies in {\em active learning}~\cite{settles2010active}.

Previous studies suggest that finding informative unlabeled samples to label is related to selecting already-labeled samples to optimize the model parameters~\cite{guillory2009active}. As reported in the previous studies~\cite{schohn2000less,bordes2005fast}, models can sometimes achieve lower generalization error after being trained with only a subset of actively selected training data. In other words, focusing on informative samples can be beneficial even when all labels are available. 

We propose two lightweight methods that actively emphasize uncertain samples to improve mini-batch SGD for classification. One method measures the variance of prediction probabilities, while the other one estimates the closeness between the prediction probabilities and the decision threshold. For logistic regression, both methods can be proven to reduce the uncertainty in the model parameters under reasonable approximations.

We present extensive experiments on CIFAR 10, CIFAR 100, MNIST (image classification), Question Type (sentence classification), CoNLL 2003, and OntoNote 5.0 (Named entity Recognition), as well as on different architectures, including multiple class logistic regression, fully-connected networks, convolutional neural networks (CNNs)~\cite{lecun1998gradient}, and residual networks~\cite{he2016deep}. The results show that active bias makes neural networks more robust without prior knowledge of noise, and reduces the generalization error by 1\% \textendash 18\% even on training sets having few (if any) annotation errors.

\section{Related work}
\label{sec:related work}
As (deep) neural networks become more widespread, many methods have recently been proposed to improve SGD training. When using (mini-batch)  SGD, the randomness of the gradient sometimes slows down the optimization, so one common approach is to use the gradient computed in previous iterations to stabilize the process. Examples include momentum~\cite{qian1999momentum}, stochastic variance reduced gradient (SVRG)~\cite{johnson2013accelerating}, and  proximal stochastic variance reduced gradient (Prox-SVRG)~\cite{xiao2014proximal}. Other work proposes variants of semi-stochastic algorithms to approximate the exact gradient direction and reduce the gradient variance~\cite{wang2013variance,mu2016stochastic}. More recently, supervised optimization methods like {\em learning by learning}~\cite{andrychowicz2016learning} also show great potential in this problem.

In addition to the high variance of the gradient, another issue with SGD is the difficulty of tuning the learning rate. Like Quasi-Newton methods, several methods adaptively adjust learning rates based on local curvature~\cite{amari2000adaptive,schaul2013no}, while ADAGRAD~\cite{duchi2011adaptive} applies different learning rates to different dimensions.  ADAM~\cite{kingma2014adam} combines several of these techniques and is widely used in practice. 

More recently, some studies accelerate SGD by weighting each class differently~\cite{gopal2016adaptive} or weighting each sample differently as we do~\cite{hinton2007recognize,ZhaoZ14,loshchilov2015online,gao2015active,alain2015variance,shrivastava2016training}, and their experiments suggest that the methods are often compatible with other techniques such as Prox-SVRG, ADAGRAD, or ADAM~\cite{loshchilov2015online,gopal2016adaptive}. Notice that Gao et al. \cite{gao2015active} discuss the idea of selecting uncertain examples for SGD based on active learning, but their proposed methods choose each sample according to the magnitude of its gradient as in ISSGD~\cite{alain2015variance}, which actually prefers more difficult examples. 

The aforementioned methods focus on accelerating the optimization of a fixed loss function given a fixed model. Many of these methods adopt importance sampling. That is, if the method prefers to select harder examples, the learning rate corresponding to those examples will be lower. This makes gradient estimation unbiased~\cite{hinton2007recognize,ZhaoZ14,alain2015variance,gao2015active,gopal2016adaptive}, which guarantees convergence~\cite{ZhaoZ14,gopal2016adaptive}.

On the other hand, to make models more robust to outliers, some approaches inject bias into the loss function in order to emphasize easier examples~\cite{pregibon1982resistant,wang2016reweighted,lee2016toward,northcutt2017learning}. Some variants of the strategy gradually increase the loss of hard examples~\cite{mandt2016variational}, as in self-paced learning~\cite{kumar2010self}. To alleviate the local minimum problem during training, other techniques that smooth the loss function have been proposed recently~\cite{chaudhari2016entropy,gulcehre2016mollifying}. Nevertheless, to our knowledge, it remains an unsolved challenge to balance the easy and difficult training examples to facilitate training while remaining robust to outliers.






\section{Methods}
\label{sec:methods}
In this section, we first discuss the baseline methods against which we shall compare and introduce some notations which we are going to use later on. We then present our two active bias methods based on prediction variance and closeness to the decision threshold.

\subsection{Baselines}

Due to its simplicity and generally good performance, the most widely used version of SGD samples each training instance uniformly. This basic strategy has two variants. The first samples with replacement. Let $\mathcal{D}={(\mathbf{x_i},y_i)}_i$ indicate the training dataset. The probability of selecting each sample is equal (i.e., $P_s(i|\mathcal{D})=\frac{1}{|\mathcal{D}|}$), so we call it {\em SGD Uniform} (SGD-Uni). The second samples without replacement. Let $S_e$ be the set of samples we have already used in the current epoch. Then, the sampling probability $P_s(i|S_e,\mathcal{D})$ would become $(\frac{1}{|\mathcal{D}|-|S_e|}) \mathbf{1}_{i \notin S_e} $, where $\mathbf{1}$ is an indicator function. This version scans through all of the examples in each epoch, so we call it {\em SGD-Scan}.


We propose a simple baseline which selects harder examples with higher probability, as done by Loshchilov and Hutter~\cite{loshchilov2015online}. Specifically, we let $P_s(i|H,S_e,\mathcal{D})\propto 1-\bar{p}_{H_i^{t-1}}(y_i|\mathbf{x_i}) + \epsilon_D$, where $H_i^{t-1}$ is the history of prediction probability which stores all $p(y_i|\mathbf{x_i})$ when $\mathbf{x_i}$ is selected to train the network before the current iteration $t$, $H=\bigcup_i H_i^{t-1}$,  $\bar{p}_{H_i^{t-1}}(y_i|\mathbf{x_i})$ is the average probability of classifying sample $i$ into its correct class $y_i$ over all the stored $p(y_i|\mathbf{x_i})$ in $H_i^{t-1}$, and $\epsilon_D$ is a smoothness constant. Notice that by only considering $p(y_i|\mathbf{x_i})$ in $H_i^{t-1}$, we won't need to perform extra forward passes. We refer to this simple baseline as {\em SGD Sampled by Difficulty} (SGD-SD).

In practice, SGD-Scan often works better than SGD-Uni because it ensures that the model sees all of the training examples in each epoch. To emphasize difficult examples while applying SGD-Scan, we weight each sample differently in the loss function. That is, the loss function is modified as $L=\sum_i v_i \cdot loss_i(W) + \lambda R(W)$, where $W$ are the parameters in the model, $loss_i(W)$ is the prediction loss, and $\lambda R(W)$ is the regularization term of the model. The weight of the $i$th sample $v_i$ can be set as $\frac{1}{N_D} (1-\bar{p}_{H_i^{t-1}}(y_i|\mathbf{x_i})+ \epsilon_D)$, where $N_D$ is a normalization constant making the average of $v_i$ equal to 1. We want to keep the average of the $v_i$  fixed so that we do not change the global learning rate. We denote this method {\em SGD Weighted by Difficulty} (SGD-WD).


Models usually cannot fit outliers well, so SGD-SD and SGD-WD would not be robust to noise. To make a model unbiased, importance sampling can be used. That is, we can let $P_s(i|H,S_e,\mathcal{D}) \propto 1-\bar{p}_{H_i^{t-1}}(y_i|\mathbf{x_i}) + \epsilon_D$ and $v_i = N_D (1-\bar{p}_{H_i^{t-1}}(y_i|\mathbf{x_i}) + \epsilon_D)^{-1}$, which is similar to an approach used by Hinton~\cite{hinton2007recognize}. We refer to this as {\em SGD Importance-Sampled by Difficulty} (SGD-ISD).

In addition, we propose two simple baselines that emphasize easy examples, as in self-paced learning. Based on the same naming convention, {\em SGD Sampled by Easiness} (SGD-SE) denotes that $P_s(i|H,S_e,\mathcal{D})\propto \bar{p}_{H_i^{t-1}}(y_i|\mathbf{x_i}) + \epsilon_E$, while {\em SGD Weighted by Easiness} (SGD-WE) sets $v_i =\frac{1}{N_E} (\bar{p}_{H_i^{t-1}}(y_i|\mathbf{x_i}) + \epsilon_E)$, where $N_E$ normalizes the $v_i$'s to have unit mean. 

\subsection{Prediction Variance}
In the active learning setting, the prediction variance can be used to measure the uncertainty of each sample for either a regression or classification problem~\cite{schein2007active}. In order to gain more information at each SGD iteration, we choose samples with high prediction variances.

Since the prediction variances are estimated on the fly, we would like to balance exploration and exploitation. Adopting the \textit{optimism in face of uncertainty} heuristics of bandit problems~\cite{bubeck2012regret}, we draw the next sample based on the estimated prediction variance plus its confidence interval. Specifically, for {\em SGD Sampled by Prediction Variance} (SGD-SPV), we let
\begin{equation}
\small
P_s(i|H,S_e,\mathcal{D}) \propto \widehat{std}^{\text{conf}}_i(H) + \epsilon_V, \text{where } \;\;\; \widehat{std}^{\text{conf}}_i(H) = \sqrt{ \widehat{var}( p_{H_i^{t-1}}(y_i|\mathbf{x_i}) )+\frac{\widehat{var}( p_{H_i^{t-1}}(y_i|\mathbf{x_i}) )^2}{|H_i^{t-1}|-1} },
\label{eq:PV}
\end{equation}
\normalsize
$\widehat{var}\left( p_{H_i^{t-1}}(y_i|\mathbf{x_i}) \right)$ is the prediction variance estimated by history $H_i^{t-1}$, and $|H_i^{t-1}|$ is the number of stored prediction probabilities. Assuming $p_{H_i^{t-1}}(y_i|\mathbf{x_i})$ is normally distributed under the uncertainty of model parameters $\mathbf{w}$, the variance of prediction variance estimation can be estimated by $2 \cdot \widehat{var}\left( p_{H_i^{t-1}}(y_i|\mathbf{x_i}) \right)^2(|H_i^{t-1}|-1)^{-1}$. 
As we did in the baselines, adding the smoothness constant $\epsilon_V $ prevents the low variance instances from never being selected again. Similarly, another variant of the method sets $v_i = \frac{1}{N_V} (\widehat{std}^{\text{conf}}_i(H) + \epsilon_V) $, where $N_V$ normalizes $v_i$ like other weighted methods; we call this {\em SGD Weighted by Prediction Variance} (SGD-WPV).

As in SGD-WD, SGD-WE or self-paced learning~\cite{avramova2015curriculum}, we train an unbiased model for several burn-in epochs at the beginning so as to judge the sampling uncertainty reasonably and stably. Other implementation details will be described in the first section of the supplementary material.

Using a low learning rate, model parameters $\mathbf{w}$ would be close to a good local minimum after sufficient burn-in epochs, and thus the posterior distribution of $\mathbf{w}$ can be locally approximated by a Gaussian distribution. Furthermore, the prediction distribution $p(y_i|\mathbf{x_i},\mathbf{w})$ is often locally smooth with respect to the model parameters $\mathbf{w}$ (i.e., small changes of model parameters only induce small changes in the prediction distribution), so a Gaussian tends to approximate the distribution of $p_{H_i^{t-1}}(y_i|\mathbf{x_i})$ well in practice.


\textbf{Example: logistic regression}

Given a Gaussian prior $Pr(W=\mathbf{w})=\mathcal{N}(\mathbf{w}|\mathbf{0},s_0 I)$ on the parameters, consider the probabilistic interpretation of logistic regression: 
\begin{equation}
\small
 -\log( Pr(Y,W=\mathbf{w}|X) )   = - \sum_i \log( p(y_i|\mathbf{x_i},\mathbf{w}) ) - \frac{c}{s_0} ||\mathbf{w}||^2,
\end{equation}
\normalsize
where $p(y_i|\mathbf{x_i},\mathbf{w})=\frac{1}{1+exp(-y_i\mathbf{w}^T\mathbf{x_i})}$, and $y_i \in \{1,-1\}$.

Since the posterior distribution of $W$ is log-concave~\cite{rennie2005regularized}, we can use $Pr(W=\mathbf{w}|Y,X)   \approx \mathcal{N}(\mathbf{w}|\mathbf{w_N},S_N)$, where $\mathbf{w_N}$ is maximum a posteriori (MAP) estimation, and
\begin{equation}
\label{eq:S_N}
\small
S_N^{-1}= \bigtriangledown_\mathbf{w} \bigtriangledown_\mathbf{w} -\log(Pr(Y,W|X))  =\sum_i p(y_i|\mathbf{x_i})\left(1-p(y_i|\mathbf{x_i})\right) \mathbf{x_i}\mathbf{x_i}^T + \frac{2c}{s_0}I.
\end{equation} 

Then, we further approximate $p(y_i|\mathbf{x_i},W)$ using the first order Taylor expansion $p(y_i|\mathbf{x_i},W)  \approx p(y_i|\mathbf{x_i},\mathbf{w})+ g_i(\mathbf{w})^T ( W-\mathbf{w} )$, where $g_i(\mathbf{w})= p(y_i|\mathbf{x_i},\mathbf{w})\left(1-p(y_i|\mathbf{x_i},\mathbf{w})\right)\mathbf{x_i}$. We can compute the prediction variance~\cite{schein2007active} with respect to the uncertainty of $W$
\begin{equation}
\label{eq:var_approx}
\small
Var( p(y_i|\mathbf{x_i},W) ) \approx g_i(\mathbf{w})^T S_N g_i(\mathbf{w}).
\end{equation}

These approximations tell us several things. First, $Var( p(y_i|\mathbf{x_i},W) )$ is proportional to $p(y_i|\mathbf{x_i},\mathbf{w})^2 (1-p(y_i|\mathbf{x_i},\mathbf{w}))^2$, so the prediction variance is larger when the sample $i$ is closer to the boundary. Second, when we have more sample points close to the boundary, the variance of the parameters $S_N$ is lower. That is, when we emphasize samples with high prediction variances, the uncertainty of parameters tends to be reduced, akin to the variance reduction strategy in active learning~\cite{mackay1992information}. Third, with a Gaussian assumption on the posterior distribution $Pr(W=\mathbf{w}|Y,X)$ and the Taylor expansion, the distribution of $p(y_i|\mathbf{x_i},W)$ in logistic regression becomes Gaussian, which justifies our previous assumption of $p_{H_i^{t-1}}(y_i|\mathbf{x_i})$ for the confidence estimation of the prediction variance. Notice that there are other methods that can measure the prediction uncertainty, such as the mutual information between labels and parameters~\cite{houlsby2011bayesian}, but we found that the prediction variance works better in our experiments.

\begin{figure*}
\centering
\captionsetup[subfigure]{justification=centering}
\begin{subfigure}{.25\textwidth}
  \centering
  \includegraphics[width=1\linewidth]{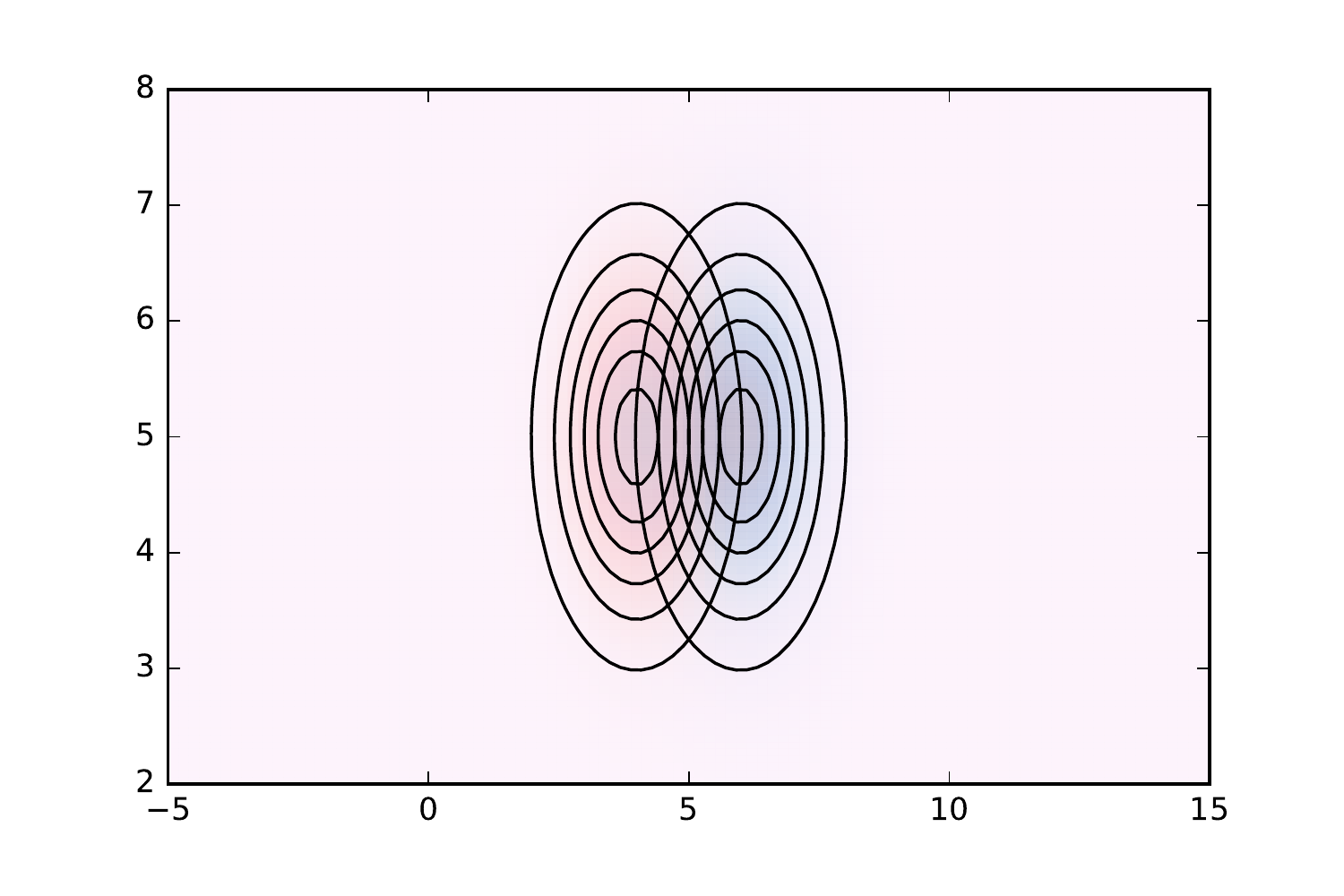}
    \vspace{-5mm}
  \caption{Sampling distribution}
  \label{fig:suba}
\end{subfigure}%
\begin{subfigure}{.25\textwidth}
  \centering
  \includegraphics[width=1\linewidth]{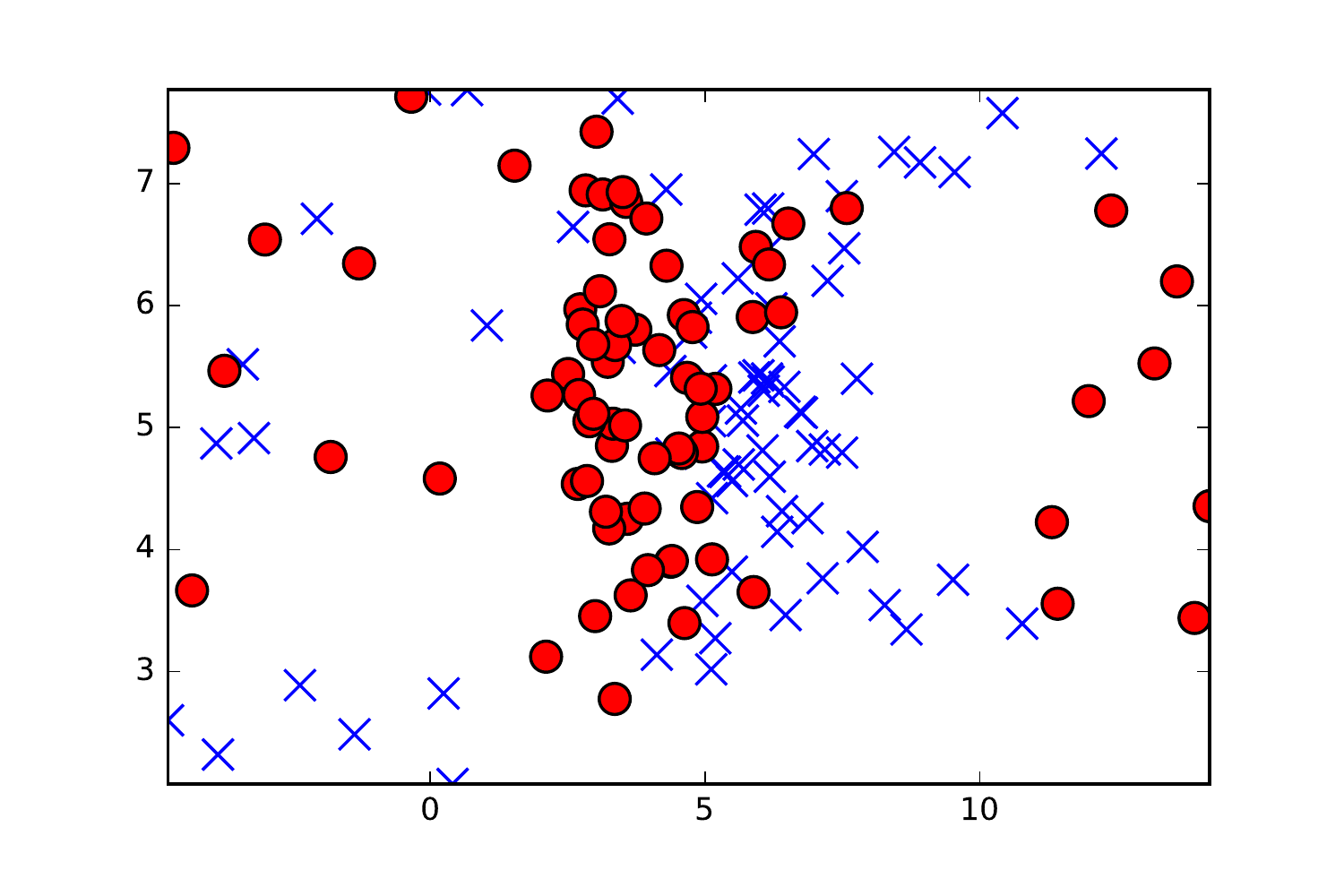}
    \vspace{-5mm}
  \caption{Training Samples}
  \label{fig:subb}
\end{subfigure}%
\begin{subfigure}{.25\textwidth}
  \centering
  \includegraphics[width=1\linewidth]{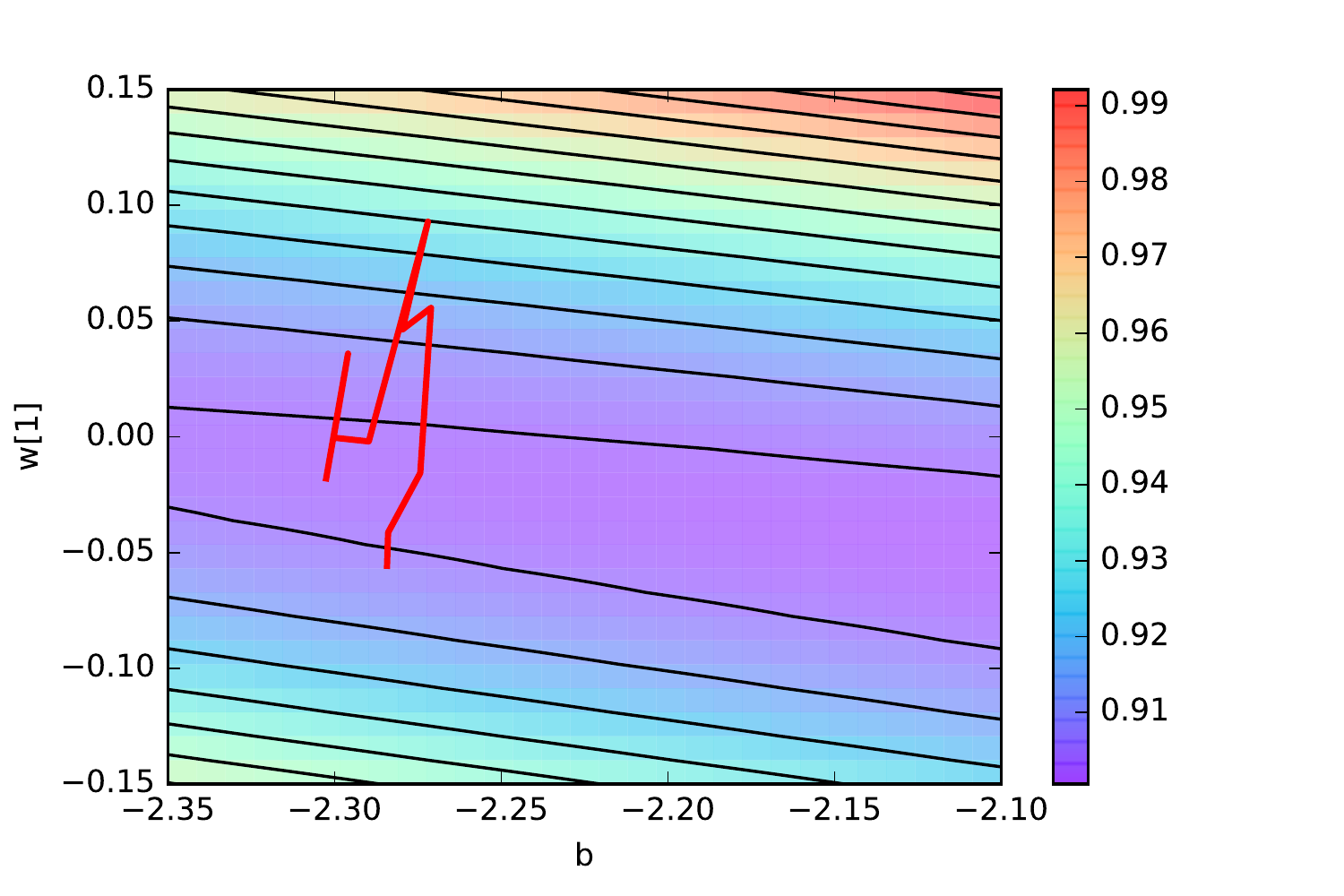}
    \vspace{-5mm}
 \caption{SGD-Scan \\ parameters space}
  \label{fig:subc}
\end{subfigure}%
\begin{subfigure}{.25\textwidth}
  \centering
  \includegraphics[width=1\linewidth]{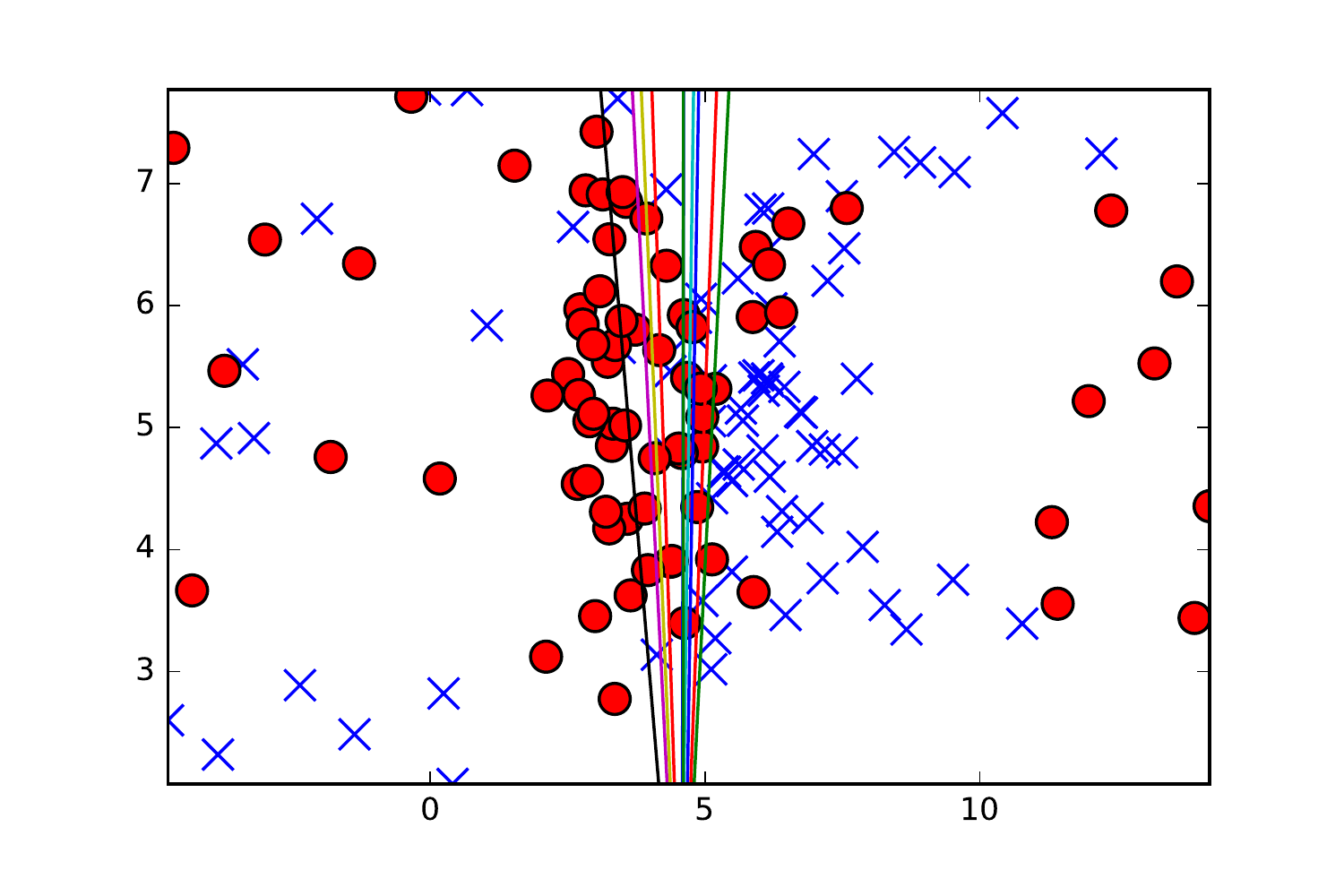}
  \vspace{-5mm}
  \caption{SGD-Scan boundaries}
  \label{fig:subd}
\end{subfigure}
\begin{subfigure}{.25\textwidth}
  \centering
  \includegraphics[width=1\linewidth]{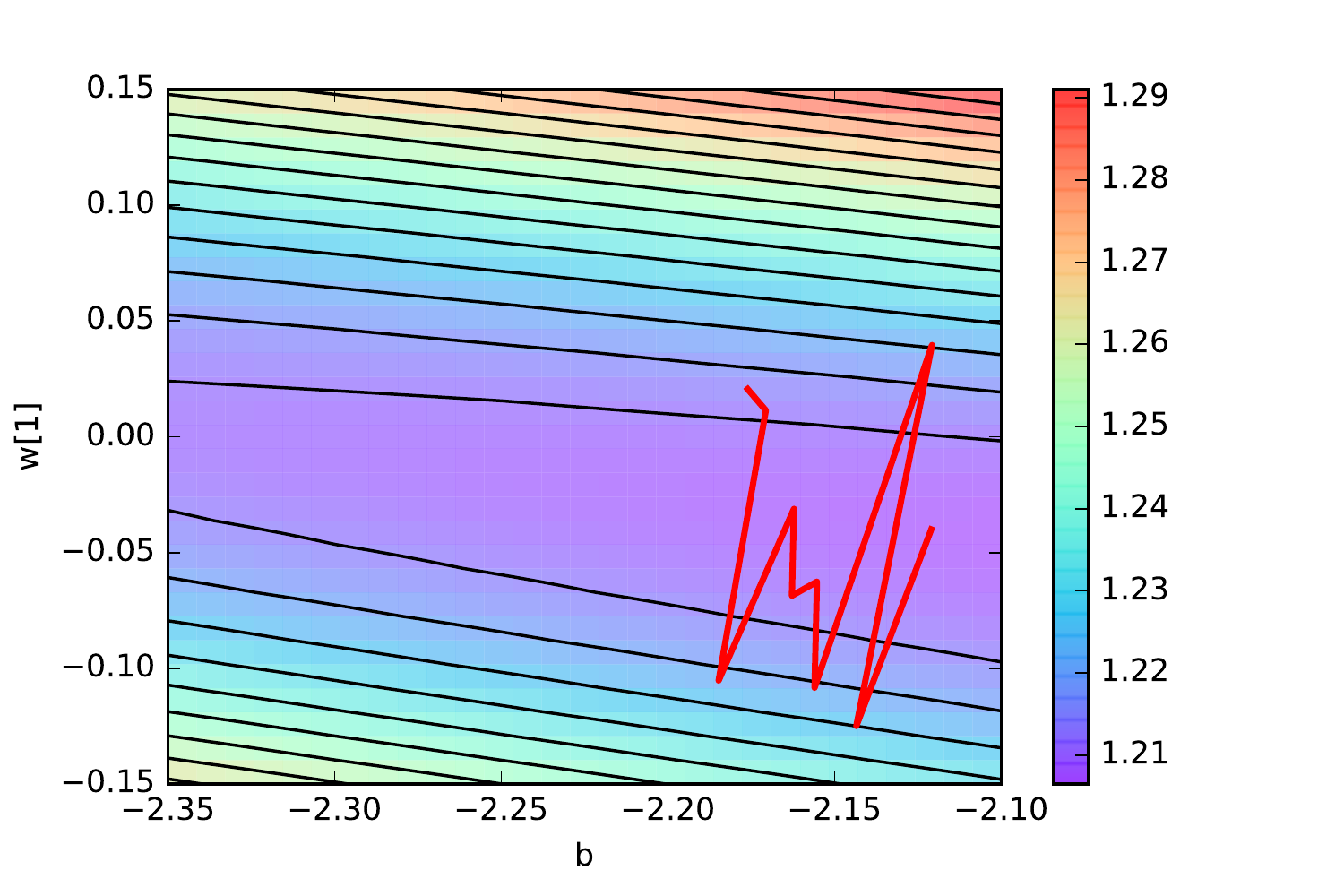}
    \vspace{-4mm}
  \caption{SGD-WD \\ parameters space}
  \label{fig:sube}
\end{subfigure}%
\begin{subfigure}{.25\textwidth}
  \centering
  \includegraphics[width=1\linewidth]{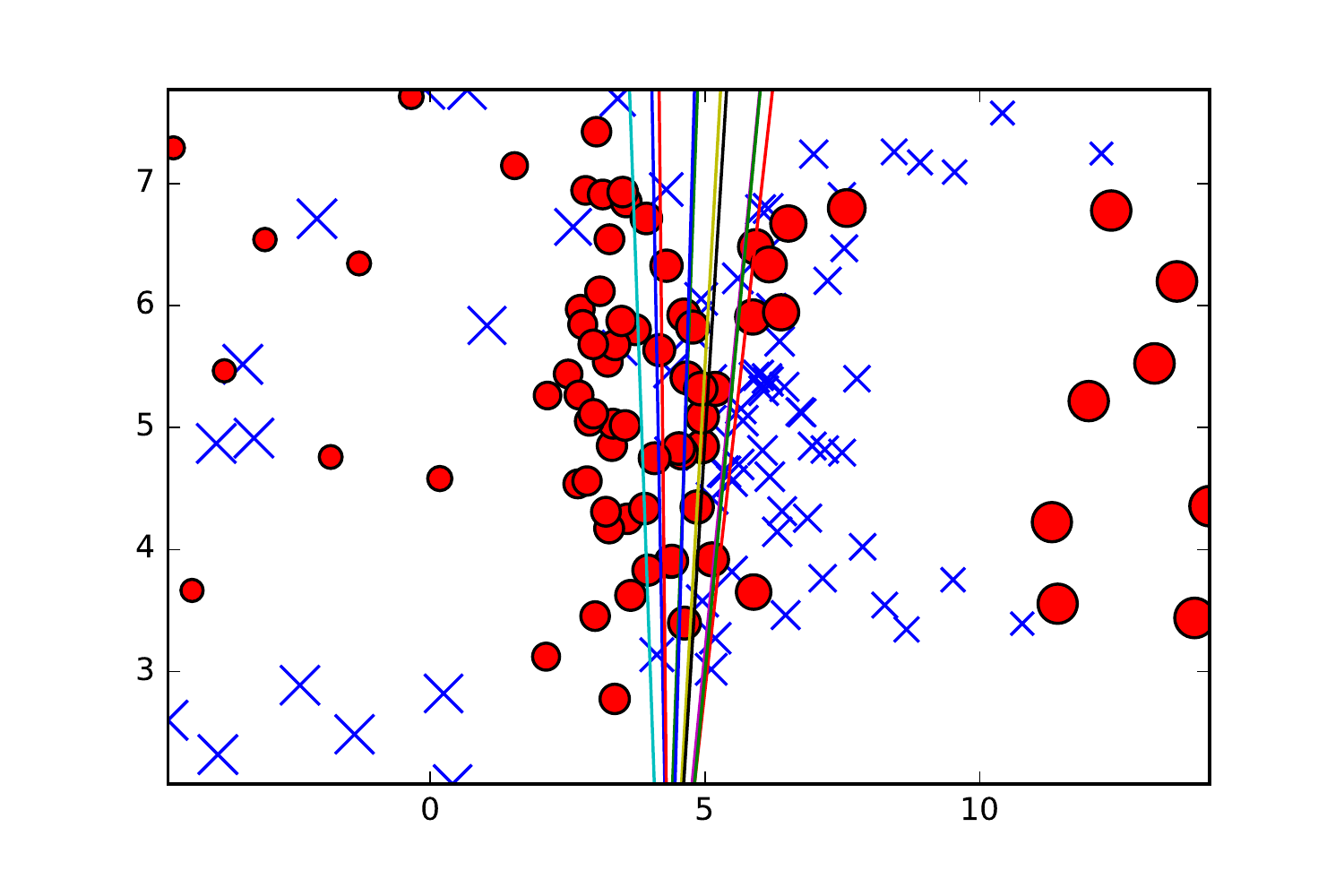}
    \vspace{-4mm}
  \caption{SGD-WD sample \\ weights and boundaries}
  \label{fig:subf}
\end{subfigure}%
\begin{subfigure}{.25\textwidth}
  \centering
  \includegraphics[width=1\linewidth]{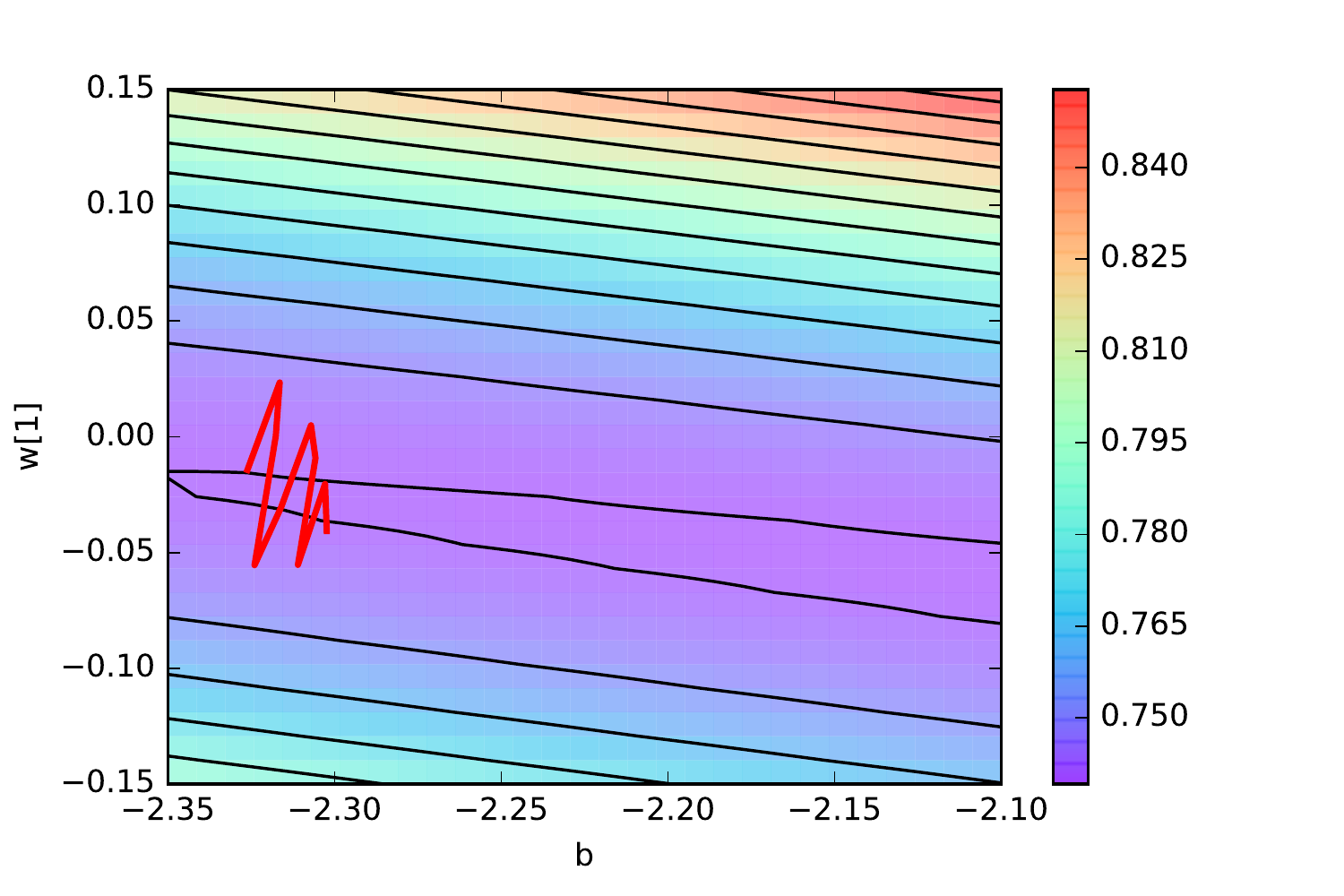}
    \vspace{-4mm}
  \caption{SGD-WPV \\ parameters space}
  \label{fig:subg}
\end{subfigure}%
\begin{subfigure}{.25\textwidth}
  \centering
  \includegraphics[width=1\linewidth]{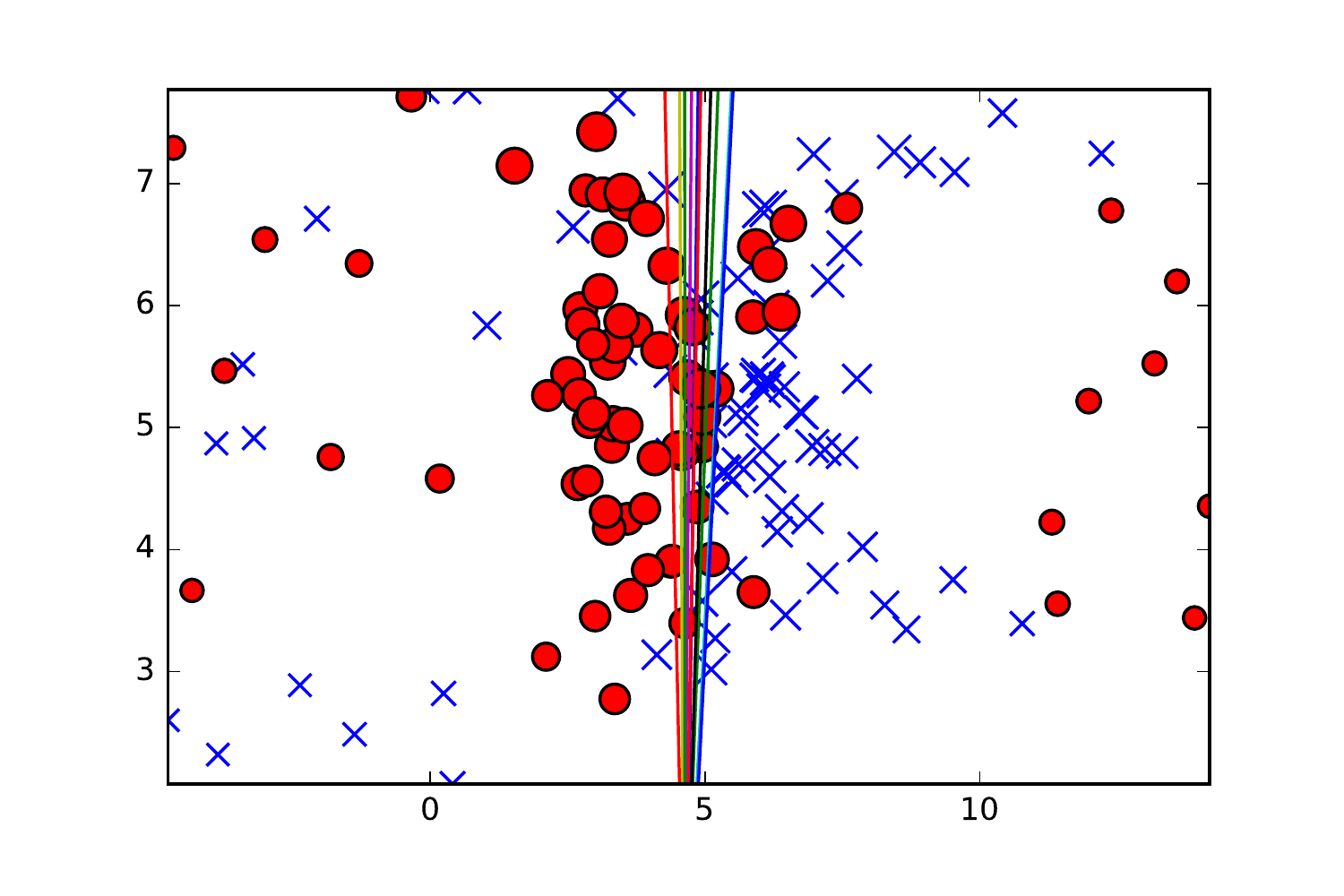}
    \vspace{-4mm}
  \caption{SGD-WPV sample \\ weights and boundaries}
  \label{fig:subh}
\end{subfigure}%
\vspace{1mm}
\caption{ A toy example which compares different methods in a two-class logistic regression model. To visualize the optimization path for the classifier parameters (the red paths in (c), (e), and (g)) in two dimensions, we fix the weight corresponding to the x-axis to 0.5 and only show the weight for y-axis $w[1]$ and bias term $b$. The $i$th sample size in (f) and (h) is proportional to $v_i$. The toy example shows that SGD-WPV can train a more accurate model in a noisy dataset. }
\vspace{-1mm}
\label{fig:toy_example}
\end{figure*}

Figure~\ref{fig:toy_example} illustrates a toy example. Given the same learning rate, we can see that the normal SGD in Figure~\ref{fig:subc} and~\ref{fig:subd} will have higher uncertainty when there are many outliers, and emphasizing difficult examples in Figure~\ref{fig:sube} and~\ref{fig:subf} makes it worse. On the other hand, the samples near the boundaries would have higher prediction variances (i.e., larger circles or crosses in Figure~\ref{fig:subh}) and thus higher impact on the loss function in SGD-WPV.

After burn-in epochs, $\textbf{w}$ becomes close to a local minimum using SGD. Then, the parameters estimated in each iteration can be viewed, approximately, as samples drawn from the posterior distribution of the parameters $Pr(W=\mathbf{w}|Y,X)$~\cite{mandt2016variational_SGD}. Therefore, after running SGD long enough, $\widehat{var}\left( p_{H_i^{t-1}}(y_i|\mathbf{x_i}) \right)$ can be used to approximate $Var\left( p(y_i|\mathbf{x_i},W) \right)$. Notice that if we directly apply bias at the beginning without running burn-in epochs, incorrect examples might be emphasized, which is also known as the local minimum problem in active learning~\cite{guillory2009active}. For instance, in Figure~\ref{fig:toy_example}, if burn-in epochs are not applied and the initial $\mathbf{w}$ is a vertical line on the left, the outliers close to the initial boundary would be emphasized, which slows down the convergence speed.

In this simple example, we can also see that the gradient magnitude is proportional to the difficulty because  $-\bigtriangledown_\mathbf{w} \log( p(y_i|\mathbf{x_i},\mathbf{w}) )=\left(1-p(y_i|\mathbf{x_i},\mathbf{w})\right)\mathbf{x_i}$. This is why we believe the SGD acceleration methods based on gradient magnitude~\cite{alain2015variance,gopal2016adaptive} can be categorized as variants of preferring difficult examples, and thus more vulnerable to outliers (like the samples on the left or right in Figure~\ref{fig:toy_example}).


\subsection{Threshold Closeness}

Motivated by the previous analysis, we propose a simpler and more direct approach to select samples whose correct class probability is close to the decision threshold. {\em SGD Sampled by Threshold Closeness} (SGD-STC) makes $P_s(i|H,S_e,\mathcal{D}) \propto \bar{p}_{H_i^{t-1}}(y_i|\mathbf{x_i})\left(1-\bar{p}_{H_i^{t-1}}(y_i|\mathbf{x_i})  \right) +\epsilon_T,$ where $\bar{p}_{H_i^{t-1}}(y_i|\mathbf{x_i})$ is the average probability of classifying sample $i$ into its correct class $y_i$ over all the stored $p(y_i|\mathbf{x_i})$ in $H_i^{t-1}$. When there are multiple classes, this measures the closeness of the threshold for distinguishing the correct class out of the union of the rest of the classes (i.e., one-versus-rest). The method is similar to an approximation of the optimal allocation in stratified sampling proposed by Druck and McCallum~\cite{druck2011toward}.

Similarly, {\em SGD Weighted by Threshold Closeness} (SGD-WTC) chooses the weight of $i$th sample $v_i=\frac{1}{N_T}\bar{p}_{H_i^{t-1}}(y_i|\mathbf{x_i})(1-\bar{p}_{H_i^{t-1}}(y_i|\mathbf{x_i}))+ \epsilon_T$, where $N_T=\frac{1}{|\mathcal{D}|}\sum_j \bar{p}_{H_j^{t-1}}(y_j|\mathbf{x_j})(1-\bar{p}_{H_j^{t-1}}(y_j|\mathbf{x_j})) + \epsilon_T$. The weighting can be viewed as combining the SGD-WD and SGD-WE by multiplying their weights together. Although other uncertainty estimates such as entropy are widely used in active learning and can also be viewed as a measure of boundary closeness, we found the proposed formula works better in our experiments.

When using logistic regression, after injecting the bias $v_i$ into the loss function, approximating the prediction probability based on previous history, removing the regularization and smoothness constant (i.e., $p(y_i|\mathbf{x_i},\mathbf{w}) \approx \bar{p}_{H_i^{t-1}}(y_i|\mathbf{x_i})$, $1/s_0=0$, and $\epsilon_T=0$), we can show that 
\footnotesize
\begin{equation}
\small
\sum_i Var( p(y_i|\mathbf{x_i},W) ) \approx \sum_i g_i(\mathbf{w})^T S_N g_i(\mathbf{w}) \approx N_T \cdot dim(w),
\label{eq:WTC}
\end{equation}
\normalsize
 where $dim(w)$ is the dimension of parameters $w$. This will ensure that the average prediction variance drops linearly as the number of training instance increases. The derivation could be seen in the supplementary materials.

\section{Experiments}
\label{sec:experiment}
\begin{table}[t]
\centering
\caption{Model architectures. Dropouts and L2 reg (regularization) are only applied to the fully-connected (FC) layer(s).}
\label{tb:exp_architecture}
\footnotesize
\begin{tabular}{ccccccccc}
\hline
\multirow{2}{*}{Dataset}  & \# Conv & Filter  & Filter   & \# Pooling & \# BN & \# FC  & Dropout & L2 \\
 &  layers &  size & number  & layers & layers & layers & keep probs & reg \\ \hline
MNIST &   2  & 5x5 & 32, 64  & 2 & 0 & 2  & 0.5 & 0.0005    \\ \hline
CIFAR 10 &  0  & N/A & N/A  & 0 & 0 & 1   &  1 & 0.01  \\ \hline
\multirow{2}{*}{CIFAR 100} &  26 or  & \multirow{2}{*}{3X3} & \multirow{2}{*}{16, 32, 64}   & \multirow{2}{*}{0} & 13 or & \multirow{2}{*}{1}  &  \multirow{2}{*}{1} & \multirow{2}{*}{0}  \\
 &  62  &  &    &  & 31 &  &   &   \\ \hline
Question Type &   1  & (2,3,4)x1 & 64 & 1 & 0 & 1   &  0.5 & 0.01 \\ \hline
CoNLL 2003  & \multirow{2}{*}{3} & \multirow{2}{*}{3x1}  & \multirow{2}{*}{100} & \multirow{2}{*}{0} & \multirow{2}{*}{0} & \multirow{2}{*}{1}  &  \multirow{2}{*}{0.5, 0.75} & \multirow{2}{*}{0.001}  \\ 
 OntoNote 5.0   &  &  &  & &  \\  \hline
 MNIST &   0  & N/A & N/A  & 0 & 0 & 2  & 1 & 0    \\ \hline
\end{tabular}
\vspace{-1mm}
\end{table}

\begin{table}[t]
\centering
\caption{Optimization hyper-parameters and experiment settings}
\label{tb:exp_setting}
\scalebox{0.9}{
\begin{tabular}{cccccccc}
\hline
\multirow{2}{*}{Dataset}  & \multirow{2}{*}{Optimizer}  & Batch & Learning & Learning  & \multirow{2}{*}{\# Epochs} & \# Burn-in & \multirow{2}{*}{\# Trials} \\
 &  & size   & rate  & rate decay  &  & epochs &  \\ \hline
MNIST & Momentum & 64 & 0.01 & 0.95  & 80  & 2 & 20 \\ \hline
CIFAR 10 & SGD & 100&  1e-6 & 0.5 (per 5 epochs)  & 30 & 10 & 30 \\ \hline
\multirow{2}{*}{CIFAR 100} & \multirow{2}{*}{Momentum} & \multirow{2}{*}{128} & \multirow{2}{*}{0.1} & 0.1 (at 80, 100,   & \multirow{2}{*}{150}  & 90 or & \multirow{2}{*}{20} \\
 & & &  & 120 epochs) &  & 50 & \\ \hline
Question Type & ADAM & 64& 0.001  & 1  & 250  & 50  & 100 \\ \hline
CoNLL 2003  & \multirow{2}{*}{ADAM}& \multirow{2}{*}{128} & \multirow{2}{*}{0.0005} & \multirow{2}{*}{1}  & \multirow{2}{*}{200} & \multirow{2}{*}{30} & \multirow{2}{*}{10} \\
 OntoNote 5.0 & & & & & & \\ \hline
 MNIST & SGD & 128 & 0.1 & 1  & 60  & 20 & 10 \\ \hline
\end{tabular}
}
\vspace{-2mm}
\end{table}

We test our methods on six different datasets. The results show that the active bias techniques constantly outperform the standard uniform sampling (i.e., SGD-Uni and SGD-Scan) in the deep models as well as the shallow models. For each dataset, we use an existing, publicly available implementation for the problem and emphasize samples using different methods. The architectures and hyper-parameters are summarized in Table~\ref{tb:exp_architecture}. All neural networks use softmax and cross-entropy loss at the last layer. The optimization and experiment setups are listed in Table~\ref{tb:exp_setting}. As shown in the second column of the table, SGD in CNNs and residual networks actually refers to momentum or ADAM instead of vanilla SGD. All experiments use mini-batch.

Like most of the widely used neural network training techniques, the proposed techniques are not applicable to every scenario. For all the datasets we tried, we found that the proposed methods are not sensitive to the hyper-parameter setup except when applying a very complicated model to a relatively smaller dataset. If a complicated model achieves 100\% training accuracy within a few epochs, the most uncertain examples would often be outliers, biasing the model towards overfitting. 

To avoid this scenario, we modify the default hyper-parameters setup in the implementation of the text classifiers in Section~\ref{sec:QT} and Section~\ref{sec:seq2seq} to achieve similar performance using simplified models. For all other models and datasets, we use the default hyper-parameters of the existing implementations, which should favor the SGD-Uni or SGD-Scan methods, since the default hyper-parameters are optimized for these cases. To show the reliability of the proposed methods, we do not optimize the hyper-parameters for the proposed methods or baselines.



Due to the randomness in all the SGD variants, we repeat experiments and list the number of trials in Table~\ref{tb:exp_setting}. At the beginning of each trial, network weights are trained with uniform sampling SGD until validation performance starts to saturate. After these burn-in epochs, we apply different sampling/weighting methods and compare performance. The number of burn-in epochs is determined by cross-validation, and the number of epochs in each trial is set large enough to let the testing error of most methods converge. In Tables~\ref{tb:sampling_results} and~\ref{tb:weight_results}, we evaluate the testing performance of each method after each epoch and report the best testing performance among epochs within each trial.

As previously discussed, there are various versions preferring easy or difficult examples. Some of them require extra time to collect necessary statistics such as the gradient magnitude of each sample~\cite{gao2015active,alain2015variance}, change the network architecture~\cite{gulcehre2016mollifying,shrivastava2016training}, or involve an annealing schedule like self-paced learning~\cite{kumar2010self,mandt2016variational}. We tried self-paced learning on CIFAR 10 but found that performance usually remains the same and is sometimes sensitive to the hyper-parameters of the annealing schedule. This finding is consistent with the results from~\cite{avramova2015curriculum}. To simplify the comparison, we focus on testing the effects of steady bias based on sample difficulty (e.g., compare with SGD-SE and SGD-SD) and do not gradually change the preference during the training like self-paced learning.


It is not always easy to change the sampling procedure because of the model or implementation constraints. For example, in sequence labeling tasks (CoNLL 2003 and OntoNote 5.0), the words in the same sentence need to be trained together. Thus, we only compare methods which modify the loss function (SGD-W*) with SGD-Scan for some models. For the other experiments, re-weighting examples (SGD-W*) generally gives us better performance than changing the sampling distribution (SGD-S*). It might be because we can better estimate the statistics of each sample.

\begin{table}[t]
\centering
\caption{The average of the best testing error rates for different sampling methods and datasets (\%). The confidence intervals are standard errors. LR means logistic regression.}
\label{tb:sampling_results}
\scalebox{0.85}{
\begin{tabular}{cccccccc}
\hline
Datasets & Model & SGD-Uni  & SGD-SD & SGD-ISD & SGD-SE &  SGD-SPV & SGD-STC \\ \hline
MNIST & CNN & 0.55$\pm$0.01&0.52$\pm$0.01 & 0.57$\pm$0.01 & 0.54$\pm$0.01 & \textbf{0.51} $\pm$0.01 & \textbf{0.51}$\pm$0.01 \\ \hline
Noisy MNIST & CNN &0.83$\pm$0.01& 1.00$\pm$0.01 & 0.84$\pm$0.01 & 0.69 $\pm$0.01 & 0.64$\pm$0.01 & \textbf{0.63}$\pm$0.01 \\ \hline
CIFAR 10 & LR &62.49$\pm$0.06&63.14$\pm$0.06 & 62.48$\pm$0.07 & 60.87$\pm$0.06 & \textbf{60.66}$\pm$0.06 & 61.00$\pm$0.06 \\ \hline
QT & CNN & 17.70$\pm$0.07&17.61$\pm$0.07&17.66$\pm$0.08&17.92$\pm$0.08&\textbf{17.49}$\pm$0.08&17.55$\pm$0.08 
\end{tabular}
}
\vspace{-1mm}
\end{table}

\begin{table}[t]
\centering
\caption{The average of the best testing error rates and their standard errors for different weighting methods (\%). For CoNLL 2003 and OntoNote 5.0, the values are 1-(F1 score). CNN, LR, RN 27, RN 63 and FC mean convolutional neural network, logistic regression, residual networks with 27 layers, residual network with 63 layers, and fully-connected network, respectively. }
\label{tb:weight_results}
\scalebox{0.85}{
\begin{tabular}{ccccccc}
\hline
Datasets & Model & SGD-Scan  & SGD-WD & SGD-WE &  SGD-WPV & SGD-WTC \\ \hline
MNIST & CNN & 0.54$\pm$0.01 & \textbf{0.48}$\pm$0.01 &  0.56$\pm$0.01 & \textbf{0.48}$\pm$0.01 & \textbf{0.48}$\pm$0.01 \\ \hline
Noisy MNIST & CNN & 0.81$\pm$0.01 & 0.92$\pm$0.01 &  0.72$\pm$0.01 & \textbf{0.61}$\pm$0.02 & 0.63$\pm$0.01 \\ \hline
CIFAR 10 & LR & 62.48$\pm$0.06 & 63.10$\pm$0.06 &  60.88$\pm$0.06 & \textbf{60.61}$\pm$0.06 & 61.02$\pm$0.06 \\ \hline
CIFAR 100 & RN 27 & 34.04$\pm$0.06 &  34.55$\pm$0.06 &  33.65$\pm$0.07 & 33.69$\pm$0.07 & \textbf{33.64}$\pm$0.07  \\ \hline
CIFAR 100 & RN 63 & 30.70$\pm$0.06 & 31.57$\pm$0.09 &  \textbf{29.92}$\pm$0.09 & 30.02$\pm$0.08 & 30.16$\pm$0.09 \\ \hline
QT & CNN & 17.79$\pm$0.08&17.70$\pm$0.08&17.87$\pm$0.08&\textbf{17.57}$\pm$0.07&17.61$\pm$0.08 \\ \hline
CoNLL 2003 & CNN & 11.62$\pm$0.04 & 11.50$\pm$0.05 & 11.73$\pm$0.04 & 11.24$\pm$0.06 & \textbf{11.18}$\pm$0.03 \\ \hline
OntoNote 5.0 & CNN & 17.80$\pm$0.05 & 17.65$\pm$0.06 & 18.40$\pm$0.05 & 17.82$\pm$0.03 & \textbf{17.51}$\pm$0.05 \\ \hline
MNIST & FC & 2.85$\pm$0.03 & \textbf{2.17}$\pm$0.01 & 3.08$\pm$0.03 & 2.68$\pm$0.02 & 2.34$\pm$0.03 \\ \hline
MNIST (distill) & FC & 2.27$\pm$0.01 & 2.13$\pm$0.02 & 2.35$\pm$0.01 & 2.18$\pm$0.02 & \textbf{2.07}$\pm$0.02 \\ \hline

\end{tabular}
}
\vspace{-1mm}
\end{table}

\subsection{MNIST}
\label{sec:exp_MNIST}
We apply our method to a CNN~\cite{lecun1998gradient} for MNIST\footnote{\small \url{http://yann.lecun.com/exdb/mnist/}} using one of the Tensorflow tutorials.\footnote{\small \url{https://github.com/tensorflow/models/blob/master/tutorials/image/mnist}} The dataset has high testing accuracy, so most of the examples are too easy for the model after a few epochs. Selecting more difficult instances can accelerate learning or improve testing accuracy~\cite{hinton2007recognize,loshchilov2015online,gopal2016adaptive}. The results from SGD-SD and SGD-WD confirm this finding while selecting uncertain examples can give us a similar or larger boost. Furthermore, we test the robustness of our methods by randomly reassigning the labels of 10\% of the images, and the results indicate that the SGD-WPV improves the performance of SGD-Scan even more while SGD-SD overfits the data seriously.

\subsection{CIFAR 10 and CIFAR 100}
We test a simple multi-class logistic regression\footnote{\small \url{https://cs231n.github.io/assignments2016/assignment2/}} on CIFAR 10~\cite{krizhevsky2009learning}.\footnote{\small \url{https://www.cs.toronto.edu/~kriz/cifar.html}} Images are down-sampled significantly to $32\times 32\times 3$, so many examples are difficult, even for humans. SGD-SPV and SGD-SE perform significantly better than SGD-Uni here, consistent with the idea that avoiding difficult examples increases robustness to outliers.

For CIFAR 100~\cite{krizhevsky2009learning}, we demonstrate that the proposed approaches can also work in very deep residual networks~\cite{he2016deep}.\footnote{\small \url{https://github.com/tensorflow/models/tree/master/resnet}} To show the method is not sensitive to the network depth and the number of burn-in epochs, we present results from the network with 27 layers and 90 burn-in epochs as well as the network with 63 layers and 50 burn-in epochs. Without changing architectures, emphasizing uncertain or easy examples gains around 0.5\% in both settings, which is significant considering the fact that the much deeper network shows only 3\% improvement here.

When training a neural network, gradually reducing the learning rate (i.e., the magnitude of gradients) usually improves performance. When difficult examples are sampled less, the magnitude of gradients would be reduced. Thus, some of the improvement of SGD-SPV and SGD-SE might come from using a lower effective learning rate. Nevertheless, since we apply the aggressive learning rate decay in the experiments of CIFAR 10 and CIFAR 100, we know that the improvements from SGD-SPV and SGD-SE cannot be entirely explained by its lower effective learning rate.


\subsection{Question Type}
\label{sec:QT}
To investigate whether our methods are effective for smaller text datasets, we apply them to a sentence classification dataset (i.e. Question Type (QT)~\cite{li2002learning}), which contains 1000 training examples and 500 testing examples.\footnote{\small \url{http://cogcomp.org/Data/QA/QC/}} We use the CNN architecture proposed by Kim~\cite{DBLP:conf/emnlp/Kim14}.\footnote{\small \url{https://github.com/dennybritz/cnn-text-classification-tf}} Like many other NLP tasks, the dataset is relatively small and this CNN classifier does not inject noise to inputs like the implementation of residual networks in CIFAR 100, so this complicated model reaches 100\% training accuracy within a few epochs.



To address this, we reduced the model complexity by 
\begin{enumerate*}[(i)]
  \item decreasing the number of filters from 128 to 64,
  \item decreasing convolutional filter widths from 3,4,5 to 2,3,4,
  \item adding L2 regularization with scale 0.01,
  \item performing PCA to reduce the dimension of pre-trained word embedding from 300 to 50 and fixing the word embedding during training.
\end{enumerate*} 
Then, the proposed active bias methods perform better than other baselines in this smaller model.






\subsection{Sequence Tagging Tasks}
\label{sec:seq2seq}
We also test our methods on Named Entity Recognition (NER) in CoNLL 2003~\cite{tjong2003introduction} and OntoNote 5.0~\cite{hovy2006ontonotes} datasets using the CNN from Strubell et al.~\cite{strubell2017fast}.\footnote{\small \url{https://github.com/iesl/dilated-cnn-ner}} Similar to Question Type, the model is too complex for our approaches. So we 
\begin{enumerate*}[(i)]
  \item only use 3 layers instead of 4 layers,
  \item reduce the number of filters from 300 to 100,
  \item add 0.001 L2 regularization,
  \item make the 50 dimension word embedding from Collobert et al.~\cite{collobert2011natural} non-trainable. 
\end{enumerate*}
The micro F1 of this smaller model only drops around 1\%-2\% from the original big model. Table~\ref{tb:weight_results} shows that our methods achieve the lowest error rate (1-F1) in both benchmarks.






\subsection{Distillation}
Although state-of-the-art neural networks in many applications memorize examples easily~\cite{zhang2016understanding}, much simpler models can usually achieve similar performance like those in the previous two experiments. In practice, such models are often preferable due to their low computation and memory requirements. We have shown that the proposed method can improve these smaller models as distillation did~\cite{hinton2015distilling}, so it is natural to check whether our methods can work well with distillation. We use an implementation\footnote{\small \url{https://github.com/akamaus/mnist-distill}} that distills a shallow CNN with 3 convolution layers to a 2 layer fully-connected network in MNIST. The teacher network can achieve 0.8\% testing error, and the temperature of softmax is set as 1.

Our approaches and baselines simply apply the sample dependent weights $v_i$ to the final loss function (i.e., cross-entropy of the true labels plus cross-entropy of the prediction probability from the teacher network). In MNIST, SGD-WTC and SGD-WD can achieve similar or better improvements compared with adding distillation into SGD-Scan. Furthermore, the best performance comes from the distillation plus SGD-WTC, which shows that active bias is compatible with distillation in this dataset.






\section{Conclusion}

Deep learning researchers often gain accuracy by employing training techniques such as momentum, dropout, batch normalization, and distillation. This paper presents a new compatible sibling to these methods, which we recommend for wide use. Our relatively simple and computationally lightweight techniques emphasize the uncertain examples (i.e., SGD-*PV and SGD-*TC). 

The experiments confirm that the proper bias can be beneficial to generalization performance. When the task is relatively easy (both training and testing accuracy are high), preferring more difficult examples works well. On the contrary, when the dataset is challenging or noisy (both training and testing accuracy are low), emphasizing easier samples often lead to a better performance. In both cases, the active bias techniques consistently lead to more accurate and robust neural networks as long as the classifier does not memorize all the training samples easily (i.e., training accuracy is high but testing accuracy is low). 


\section*{Acknowledgements} 
This material is based on research sponsored by National Science Foundation under Grant No. 1514053 and by DARPA under agreement number FA8750-1 3-2-0020 and HRO011-15-2-0036. The U.S. Government is authorized to reproduce and distribute reprints for Governmental purposes notwithstanding any copyright notation thereon. The views and conclusions contained herein are those of the authors and should not be interpreted as necessarily representing the official policies or endorsements, either expressed or implied, of DARPA or the U.S. Government.

\bibliographystyle{abbrvnat}
\bibliography{example_paper}

\begin{thebibliography}{51}
\providecommand{\natexlab}[1]{#1}
\providecommand{\url}[1]{\texttt{#1}}
\expandafter\ifx\csname urlstyle\endcsname\relax
  \providecommand{\doi}[1]{doi: #1}\else
  \providecommand{\doi}{doi: \begingroup \urlstyle{rm}\Url}\fi

\bibitem[Alain et~al.(2015)Alain, Lamb, Sankar, Courville, and
  Bengio]{alain2015variance}
G.~Alain, A.~Lamb, C.~Sankar, A.~Courville, and Y.~Bengio.
\newblock Variance reduction in {SGD} by distributed importance sampling.
\newblock \emph{arXiv preprint arXiv:1511.06481}, 2015.

\bibitem[Amari et~al.(2000)Amari, Park, and Fukumizu]{amari2000adaptive}
S.-I. Amari, H.~Park, and K.~Fukumizu.
\newblock Adaptive method of realizing natural gradient learning for multilayer
  perceptrons.
\newblock \emph{Neural Computation}, 12\penalty0 (6):\penalty0 1399--1409,
  2000.

\bibitem[Andrychowicz et~al.(2016)Andrychowicz, Denil, Gomez, Hoffman, Pfau,
  Schaul, and de~Freitas]{andrychowicz2016learning}
M.~Andrychowicz, M.~Denil, S.~Gomez, M.~W. Hoffman, D.~Pfau, T.~Schaul, and
  N.~de~Freitas.
\newblock Learning to learn by gradient descent by gradient descent.
\newblock In \emph{NIPS}, 2016.

\bibitem[Avramova(2015)]{avramova2015curriculum}
V.~Avramova.
\newblock Curriculum learning with deep convolutional neural networks, 2015.

\bibitem[Bengio et~al.(2009)Bengio, Louradour, Collobert, and
  Weston]{bengio2009curriculum}
Y.~Bengio, J.~Louradour, R.~Collobert, and J.~Weston.
\newblock Curriculum learning.
\newblock In \emph{ICML}, 2009.

\bibitem[Bordes et~al.(2005)Bordes, Ertekin, Weston, and
  Bottou]{bordes2005fast}
A.~Bordes, S.~Ertekin, J.~Weston, and L.~Bottou.
\newblock Fast kernel classifiers with online and active learning.
\newblock \emph{Journal of Machine Learning Research}, 6\penalty0
  (Sep):\penalty0 1579--1619, 2005.

\bibitem[Bubeck et~al.(2012)Bubeck, Cesa-Bianchi, et~al.]{bubeck2012regret}
S.~Bubeck, N.~Cesa-Bianchi, et~al.
\newblock Regret analysis of stochastic and nonstochastic multi-armed bandit
  problems.
\newblock \emph{Foundations and Trends{\textregistered} in Machine Learning},
  5\penalty0 (1):\penalty0 1--122, 2012.

\bibitem[Chaudhari et~al.(2017)Chaudhari, Choromanska, Soatto, and
  LeCun]{chaudhari2016entropy}
P.~Chaudhari, A.~Choromanska, S.~Soatto, and Y.~LeCun.
\newblock {Entropy-SGD}: Biasing gradient descent into wide valleys.
\newblock In \emph{ICLR}, 2017.

\bibitem[Collobert et~al.(2011)Collobert, Weston, Bottou, Karlen, Kavukcuoglu,
  and Kuksa]{collobert2011natural}
R.~Collobert, J.~Weston, L.~Bottou, M.~Karlen, K.~Kavukcuoglu, and P.~Kuksa.
\newblock Natural language processing (almost) from scratch.
\newblock \emph{Journal of Machine Learning Research}, 12\penalty0
  (Aug):\penalty0 2493--2537, 2011.

\bibitem[Druck and McCallum(2011)]{druck2011toward}
G.~Druck and A.~McCallum.
\newblock Toward interactive training and evaluation.
\newblock In \emph{Proceedings of the 20th ACM international conference on
  Information and knowledge management}, pages 947--956. ACM, 2011.

\bibitem[Duchi et~al.(2011)Duchi, Hazan, and Singer]{duchi2011adaptive}
J.~Duchi, E.~Hazan, and Y.~Singer.
\newblock Adaptive subgradient methods for online learning and stochastic
  optimization.
\newblock \emph{Journal of Machine Learning Research}, 12\penalty0
  (Jul):\penalty0 2121--2159, 2011.

\bibitem[Gao et~al.(2015)Gao, Jagadish, and Ooi]{gao2015active}
J.~Gao, H.~Jagadish, and B.~C. Ooi.
\newblock Active sampler: Light-weight accelerator for complex data analytics
  at scale.
\newblock \emph{arXiv preprint arXiv:1512.03880}, 2015.

\bibitem[Gopal(2016)]{gopal2016adaptive}
S.~Gopal.
\newblock Adaptive sampling for {SGD} by exploiting side information.
\newblock In \emph{ICML}, 2016.

\bibitem[Guillory et~al.(2009)Guillory, Chastain, and
  Bilmes]{guillory2009active}
A.~Guillory, E.~Chastain, and J.~A. Bilmes.
\newblock Active learning as non-convex optimization.
\newblock In \emph{AISTATS}, 2009.

\bibitem[Gulcehre et~al.(2017)Gulcehre, Moczulski, Visin, and
  Bengio]{gulcehre2016mollifying}
C.~Gulcehre, M.~Moczulski, F.~Visin, and Y.~Bengio.
\newblock Mollifying networks.
\newblock In \emph{ICLR}, 2017.

\bibitem[He et~al.(2016)He, Zhang, Ren, and Sun]{he2016deep}
K.~He, X.~Zhang, S.~Ren, and J.~Sun.
\newblock Deep residual learning for image recognition.
\newblock In \emph{Proceedings of the IEEE Conference on Computer Vision and
  Pattern Recognition}, pages 770--778, 2016.

\bibitem[Hinton et~al.(2014)Hinton, Vinyals, and Dean]{hinton2015distilling}
G.~Hinton, O.~Vinyals, and J.~Dean.
\newblock Distilling the knowledge in a neural network.
\newblock In \emph{NIPS Deep Learning Workshop}, 2014.

\bibitem[Hinton(2007)]{hinton2007recognize}
G.~E. Hinton.
\newblock To recognize shapes, first learn to generate images.
\newblock \emph{Progress in brain research}, 165:\penalty0 535--547, 2007.

\bibitem[Houlsby et~al.(2011)Houlsby, Husz{\'a}r, Ghahramani, and
  Lengyel]{houlsby2011bayesian}
N.~Houlsby, F.~Husz{\'a}r, Z.~Ghahramani, and M.~Lengyel.
\newblock Bayesian active learning for classification and preference learning.
\newblock \emph{arXiv preprint arXiv:1112.5745}, 2011.

\bibitem[Hovy et~al.(2006)Hovy, Marcus, Palmer, Ramshaw, and
  Weischedel]{hovy2006ontonotes}
E.~Hovy, M.~Marcus, M.~Palmer, L.~Ramshaw, and R.~Weischedel.
\newblock {OntoNotes}: the 90\% solution.
\newblock In \emph{HLT-NAACL}, 2006.

\bibitem[Johnson and Zhang(2013)]{johnson2013accelerating}
R.~Johnson and T.~Zhang.
\newblock Accelerating stochastic gradient descent using predictive variance
  reduction.
\newblock In \emph{NIPS}, 2013.

\bibitem[Kim(2014)]{DBLP:conf/emnlp/Kim14}
Y.~Kim.
\newblock Convolutional neural networks for sentence classification.
\newblock In \emph{EMNLP}, 2014.

\bibitem[Kingma and Ba(2014)]{kingma2014adam}
D.~Kingma and J.~Ba.
\newblock Adam: A method for stochastic optimization.
\newblock \emph{arXiv preprint arXiv:1412.6980}, 2014.

\bibitem[Krizhevsky and Hinton(2009)]{krizhevsky2009learning}
A.~Krizhevsky and G.~Hinton.
\newblock Learning multiple layers of features from tiny images.
\newblock 2009.

\bibitem[Kumar et~al.(2010)Kumar, Packer, and Koller]{kumar2010self}
M.~P. Kumar, B.~Packer, and D.~Koller.
\newblock Self-paced learning for latent variable models.
\newblock In \emph{NIPS}, 2010.

\bibitem[LeCun et~al.(1998)LeCun, Bottou, Bengio, and
  Haffner]{lecun1998gradient}
Y.~LeCun, L.~Bottou, Y.~Bengio, and P.~Haffner.
\newblock Gradient-based learning applied to document recognition.
\newblock \emph{Proceedings of the IEEE}, 86\penalty0 (11):\penalty0
  2278--2324, 1998.

\bibitem[Lee et~al.(2016)Lee, Yang, and Lin]{lee2016toward}
G.-H. Lee, S.-W. Yang, and S.-D. Lin.
\newblock Toward implicit sample noise modeling: Deviation-driven matrix
  factorization.
\newblock \emph{arXiv preprint arXiv:1610.09274}, 2016.

\bibitem[Li and Roth(2002)]{li2002learning}
X.~Li and D.~Roth.
\newblock Learning question classifiers.
\newblock In \emph{COLING}, 2002.

\bibitem[Loshchilov and Hutter(2015)]{loshchilov2015online}
I.~Loshchilov and F.~Hutter.
\newblock Online batch selection for faster training of neural networks.
\newblock \emph{arXiv preprint arXiv:1511.06343}, 2015.

\bibitem[MacKay(1992)]{mackay1992information}
D.~J. MacKay.
\newblock Information-based objective functions for active data selection.
\newblock \emph{Neural computation}, 4\penalty0 (4):\penalty0 590--604, 1992.

\bibitem[Mandt et~al.(2016{\natexlab{a}})Mandt, Hoffman, and
  Blei]{mandt2016variational_SGD}
S.~Mandt, M.~D. Hoffman, and D.~M. Blei.
\newblock A variational analysis of stochastic gradient algorithms.
\newblock In \emph{ICML}, 2016{\natexlab{a}}.

\bibitem[Mandt et~al.(2016{\natexlab{b}})Mandt, McInerney, Abrol, Ranganath,
  and Blei]{mandt2016variational}
S.~Mandt, J.~McInerney, F.~Abrol, R.~Ranganath, and D.~Blei.
\newblock Variational tempering.
\newblock In \emph{AISTATS}, 2016{\natexlab{b}}.

\bibitem[Meng et~al.(2015)Meng, Zhao, and Jiang]{meng2015objective}
D.~Meng, Q.~Zhao, and L.~Jiang.
\newblock What objective does self-paced learning indeed optimize?
\newblock \emph{arXiv preprint arXiv:1511.06049}, 2015.

\bibitem[Mu et~al.(2016)Mu, Liu, Liu, and Fan]{mu2016stochastic}
Y.~Mu, W.~Liu, X.~Liu, and W.~Fan.
\newblock Stochastic gradient made stable: A manifold propagation approach for
  large-scale optimization.
\newblock \emph{IEEE Transactions on Knowledge and Data Engineering}, 2016.

\bibitem[Northcutt et~al.(2017)Northcutt, Wu, and
  Chuang]{northcutt2017learning}
C.~G. Northcutt, T.~Wu, and I.~L. Chuang.
\newblock Learning with confident examples: Rank pruning for robust
  classification with noisy labels.
\newblock \emph{arXiv preprint arXiv:1705.01936}, 2017.

\bibitem[Pi et~al.(2016)Pi, Li, Zhang, Meng, Wu, Xiao, and Zhuang]{pi2016self}
T.~Pi, X.~Li, Z.~Zhang, D.~Meng, F.~Wu, J.~Xiao, and Y.~Zhuang.
\newblock Self-paced boost learning for classification.
\newblock In \emph{IJCAI}, 2016.

\bibitem[Pregibon(1982)]{pregibon1982resistant}
D.~Pregibon.
\newblock Resistant fits for some commonly used logistic models with medical
  applications.
\newblock \emph{Biometrics}, pages 485--498, 1982.

\bibitem[Qian(1999)]{qian1999momentum}
N.~Qian.
\newblock On the momentum term in gradient descent learning algorithms.
\newblock \emph{Neural networks}, 12\penalty0 (1):\penalty0 145--151, 1999.

\bibitem[Rennie(2005)]{rennie2005regularized}
J.~D. Rennie.
\newblock Regularized logistic regression is strictly convex.
\newblock \emph{Unpublished manuscript. URL:
  \url{people.csail.mit.edu/jrennie/writing/convexLR.pdf}}, 2005.

\bibitem[Schaul et~al.(2013)Schaul, Zhang, and LeCun]{schaul2013no}
T.~Schaul, S.~Zhang, and Y.~LeCun.
\newblock No more pesky learning rates.
\newblock \emph{ICML}, 2013.

\bibitem[Schein and Ungar(2007)]{schein2007active}
A.~I. Schein and L.~H. Ungar.
\newblock Active learning for logistic regression: an evaluation.
\newblock \emph{Machine Learning}, 68\penalty0 (3):\penalty0 235--265, 2007.

\bibitem[Schohn and Cohn(2000)]{schohn2000less}
G.~Schohn and D.~Cohn.
\newblock Less is more: Active learning with support vector machines.
\newblock In \emph{ICML}, 2000.

\bibitem[Settles(2010)]{settles2010active}
B.~Settles.
\newblock Active learning literature survey.
\newblock \emph{University of Wisconsin, Madison}, 52\penalty0
  (55-66):\penalty0 11, 2010.

\bibitem[Shrivastava et~al.(2016)Shrivastava, Gupta, and
  Girshick]{shrivastava2016training}
A.~Shrivastava, A.~Gupta, and R.~Girshick.
\newblock Training region-based object detectors with online hard example
  mining.
\newblock In \emph{CVPR}, 2016.

\bibitem[Strubell et~al.(2017)Strubell, Verga, Belanger, and
  McCallum]{strubell2017fast}
E.~Strubell, P.~Verga, D.~Belanger, and A.~McCallum.
\newblock Fast and accurate sequence labeling with iterated dilated
  convolutions.
\newblock \emph{arXiv preprint arXiv:1702.02098}, 2017.

\bibitem[Tjong Kim~Sang and De~Meulder(2003)]{tjong2003introduction}
E.~F. Tjong Kim~Sang and F.~De~Meulder.
\newblock Introduction to the conll-2003 shared task: Language-independent
  named entity recognition.
\newblock In \emph{HLT-NAACL}, 2003.

\bibitem[Wang et~al.(2013)Wang, Chen, Smola, and Xing]{wang2013variance}
C.~Wang, X.~Chen, A.~J. Smola, and E.~P. Xing.
\newblock Variance reduction for stochastic gradient optimization.
\newblock In \emph{NIPS}, 2013.

\bibitem[Wang et~al.(2016)Wang, Kucukelbir, and Blei]{wang2016reweighted}
Y.~Wang, A.~Kucukelbir, and D.~M. Blei.
\newblock Reweighted data for robust probabilistic models.
\newblock \emph{arXiv preprint arXiv:1606.03860}, 2016.

\bibitem[Xiao and Zhang(2014)]{xiao2014proximal}
L.~Xiao and T.~Zhang.
\newblock A proximal stochastic gradient method with progressive variance
  reduction.
\newblock \emph{SIAM Journal on Optimization}, 24\penalty0 (4):\penalty0
  2057--2075, 2014.

\bibitem[Zhang et~al.(2017)Zhang, Bengio, Hardt, Recht, and
  Vinyals]{zhang2016understanding}
C.~Zhang, S.~Bengio, M.~Hardt, B.~Recht, and O.~Vinyals.
\newblock Understanding deep learning requires rethinking generalization.
\newblock In \emph{ICLR}, 2017.

\bibitem[Zhao and Zhang(2014)]{ZhaoZ14}
P.~Zhao and T.~Zhang.
\newblock Stochastic optimization with importance sampling.
\newblock \emph{arXiv preprint arXiv:1412.2753}, 2014.

\end{thebibliography}

\newpage

\appendix
\renewcommand{\algorithmicrequire}{\textbf{Input:}}
\renewcommand{\algorithmicensure}{\textbf{Output:}}

\begin{algorithm}[h]
    \caption{SGD Training with Sample Emphasis}
    \label{alg1}
    \begin{algorithmic}

	\REQUIRE{Training data $\mathcal{D}$, Batch size $|B|$, Number of class $|C|$, \# epochs $E$, \# burn-in epochs $e_b$}
	\ENSURE{NN}
    \STATE $ $Initialize all weights $W$ in NN$ $
    \STATE $ H_i \gets \{\frac{1}{|C|} \}$ for all training sample $i$
    \STATE $ v_i \gets 1 $ for all training sample $i$
    \STATE $ t \gets 1 $
    \FOR{epoch $e \gets 1 ... E$}
    	    \STATE $S_e \gets \emptyset$
        \FOR{each iteration}
			\IF {$e > e_b$}
				   \STATE $ $Sample $B$ according to $P_s(i|H,S_e,\mathcal{D})$
            \ELSE
           		   \STATE $ $Sample $B$ uniformly from $\mathcal{D}$ 
              \ENDIF
            \STATE $ $Weight sample $i$ by $v_i$ for all $i$ in $B$
            \STATE $ $Update parameters $W$ in NN $ $
            \FOR{$i$ in $B$}
				\STATE $H_i  \gets H_i \cup \{p_t(y_i|\mathbf{x_i}) \}$
                \STATE $S_e \gets S_e \cup \{i\}$
				\STATE $ $Update $P_s(i|H,S_e,\mathcal{D})$ and $v_i$.
            \ENDFOR
            \STATE $t \gets t+1$
        \ENDFOR
      \ENDFOR
\end{algorithmic}
\end{algorithm}

\section{Implementation details}
\label{sec:implementation}
The general framework of the methods can be seen in Algorithm~\ref{alg1}. In each aforementioned method, if $P_s(i|H,S_e,\mathcal{D})$ is not specified, we use SGD-Scan (without replacement uniform sampling). If $v_i$ is not specified, it means $v_i=1$ for all sample $i$. The $\epsilon$ in each method is set as the average of current estimation. For example, $\epsilon_D$ for SDG Sampled by Difficulty is set as $\frac{1}{|\mathcal{D}|}\sum_j  1-\bar{p}_{H_j^{t-1}}(y_j|\mathbf{x_j})$.

When estimating sample related statistics like prediction variance, we found that excluding the prediction history near the beginning transient state improves performance. In our implementation, we use a simple outlier removal by computing the deviation between the prediction probability and its average at iteration $t$ (i.e., $d_t=\left|p(y_i|\mathbf{x_i})-\bar{p}_{H_i^{t-1}}(y_i|\mathbf{x_i})\right|$), and excludes all prediction probability $p_t(y_i|\mathbf{x_i})$ at current iteration when $d_t >2 \cdot median_{t'}(d_{t'})$. We apply the same method when estimating difficulty, easiness, prediction variance or threshold closeness.

By only using the prediction results from previous iterations, implementing the methods is easy and the overhead of the method is very small because we do not need any extra forward or backward passes in the neural network. Due to the outlier removal process, the average overhead for each sample at each epoch is $O(E)$, where $E$ is the number of total epochs. 

When we have a very large number of samples and epochs, we can modify outlier removal by only considering the prediction probability in the latest few epochs. Then, the overhead is constant. In Section~\ref{sec:seq2seq}, performing outlier removal in the prediction history of each word is time-consuming, so we determine the uncertainty only based on the latest 5 epochs.

\newpage

\section{Experiment details}
\label{sec:experiments}
Summaries of dataset properties can be seen in Table~\ref{tb:dataset_stats}. 

\begin{figure*}[h!]
\centering
\begin{minipage}{.5\textwidth}
  \centering
  \includegraphics[width=1\linewidth]{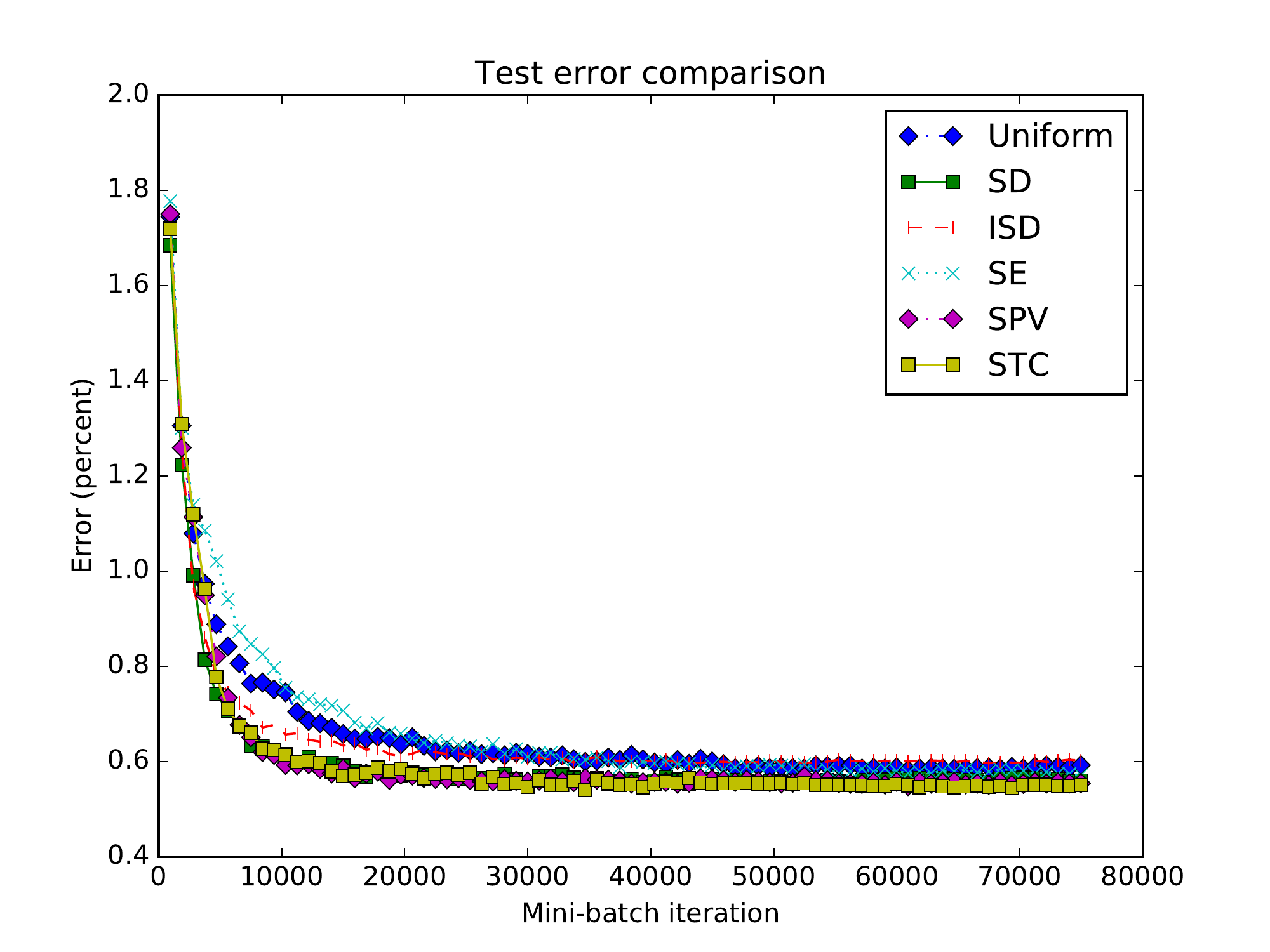}
  \caption{MNIST error rate (\%)}
  \label{fig:MNIST_sampling_method}
\end{minipage}%
\begin{minipage}{.5\textwidth}
  \centering
  \includegraphics[width=1\linewidth]{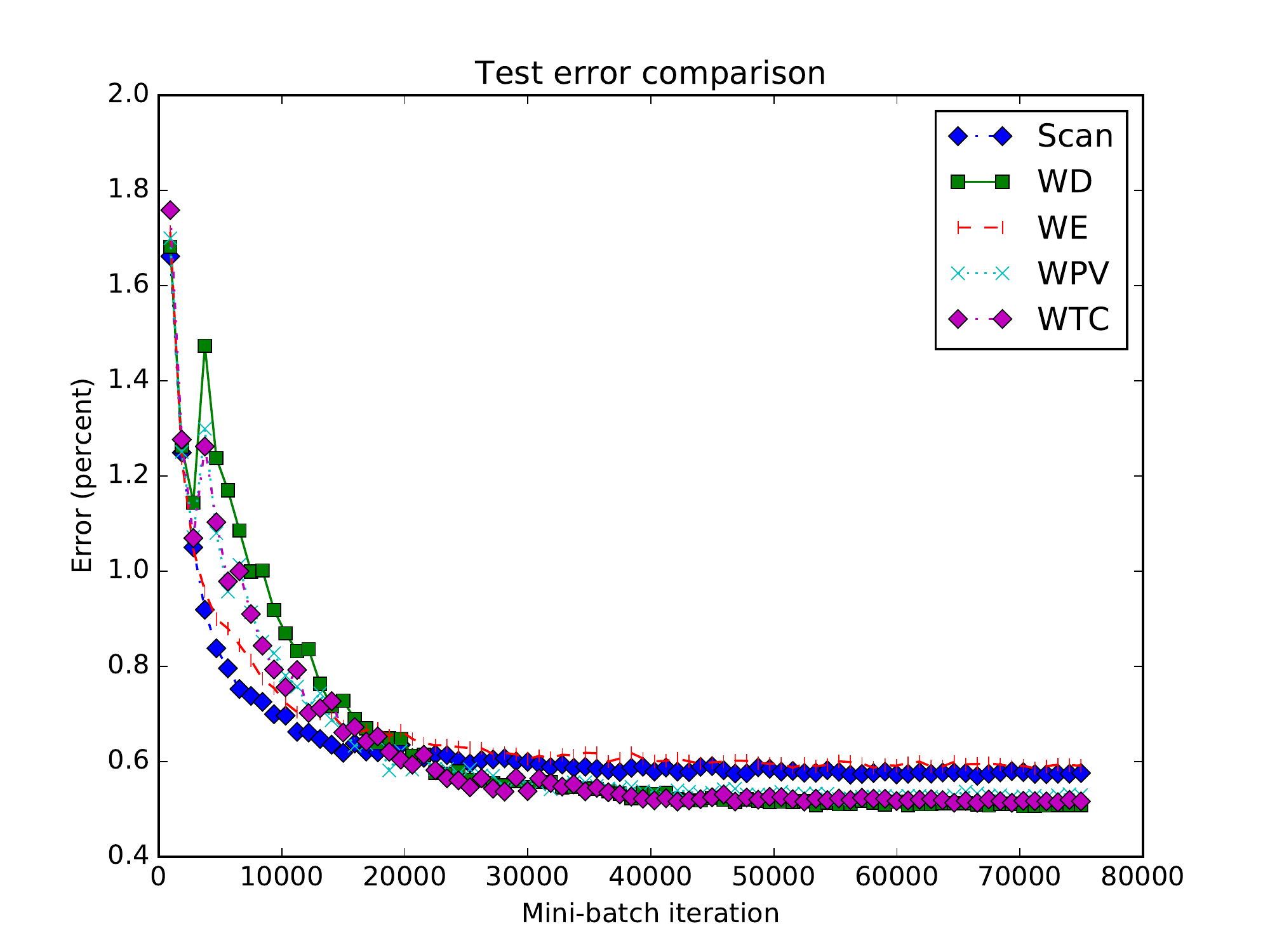}
  \caption{MNIST error rate (\%)}
  \label{fig:MNIST_weighted_method}
\end{minipage}
\end{figure*}

In Figure~\ref{fig:MNIST_sampling_method} and Figure~\ref{fig:MNIST_weighted_method}, we present the convergence curves of MNIST without noise for the experiment in Section~\ref{sec:exp_MNIST}. By comparing the error rates, we can see that changing the sampling distribution accelerates the training more, but changing the loss function can give us better results at the end.

In the paper, we only provide the best testing performance within each trial. To further understand the characteristics of each method, we report the average testing performance of the last 10 epochs in Table~\ref{tb:sampling_avg_results} and Table~\ref{tb:weighting_avg_results}. The results in the tables roughly follow the same trends in Table~\ref{tb:sampling_results} and Table~\ref{tb:weight_results}. In addition, the training errors are presented in Table~\ref{tb:sampling_train_results} and Table~\ref{tb:weighting_train_results}. We can see that emphasizing difficult examples indeed usually increases the training accuracy, but it does not necessarily imply the improvements in the testing error.

\newpage

\begin{table*}[t]
\centering
\caption{Dataset Statistics. The average sentence length $L$ in Question Type, CoNLL 2003, and OntoNote 5.0 datasets are 11, 14, and 18, respectively. CoNLL 2003 and OntoNote 5.0 are sequence tagging task, so each word is an instance with a label.}
\label{tb:dataset_stats}
\begin{tabular}{|c|c|c|c|c|c|}
\hline
Dataset  & \# Class  & Instance & Input dimensions  & \# Training  & \# Testing  \\ \hline
MNIST & 10 & Image & 28x28 & 60,000  & 10,000  \\
CIFAR 10 & 10  & Image & 32x32x3 & 50,000 & 10,000  \\
CIFAR 100 & 100  & Image & 32x32x3 & 50,000 & 10,000  \\
Question Type & 6  & Sentence & 50 $\times L$  & 1000 & 500 \\
CoNLL 2003 & 17  & Word & 50 $\times L$ & 204,567  & 46,666	 \\
OntoNote 5.0  & 74  & Word & 50 $\times L$ & 1,088,503  & 152,728 \\ \hline
\end{tabular}
\end{table*}

\begin{table}[h]
\centering
\caption{Testing error rates (last 10 epochs) of sampling methods (\%). Notice that we drop the whole rows of standard errors in the table when they are all below 0.01\%.}
\label{tb:sampling_avg_results}
\scalebox{0.8}{
\begin{tabular}{cccccccc}
\hline
Datasets & Model & SGD-Uni  & SGD-SD & SGD-ISD & SGD-SE &  SGD-SPV & SGD-STC \\ \hline
MNIST & CNN & 0.59 & 0.56 & 0.60 & 0.58 & \textbf{0.55} & \textbf{0.55} \\ \hline
Noisy MNIST & CNN & 1.18$\pm$0.00 & 1.52$\pm$0.01 & 1.26$\pm$0.00 & \textbf{0.76}$\pm$0.00 & 0.92$\pm$0.00 & 0.85$\pm$0.00 \\ \hline
CIFAR 10 & LR & 62.66$\pm$0.01 & 63.35$\pm$0.01 & 62.64$\pm$0.01 & 61.01$\pm$0.01 & \textbf{60.80}$\pm$0.01 & 61.16$\pm$0.01 \\ \hline
QT & CNN & 18.56$\pm$0.01&18.48$\pm$0.01&18.51$\pm$0.01&18.79$\pm$0.01&\textbf{18.33}$\pm$0.01&18.44$\pm$0.01 \\ \hline
\end{tabular}
}
\end{table}

\begin{table}[h]
\centering
\caption{Testing error rates (last 10 epochs) of sampling methods (\%). For CoNLL 2003 and OntoNote 5.0, the values are 1-(F1 score). When all standard errors in a row are smaller than 0.01, we skip them in the table. }
\label{tb:weighting_avg_results}
\scalebox{0.8}{
\begin{tabular}{ccccccc}
\hline
Datasets & Model & SGD-Scan  & SGD-WD & SGD-WE &  SGD-WPV & SGD-WTC \\ \hline
MNIST & CNN & 0.58 & \textbf{0.51} & 0.59 & 0.53 & 0.52 \\ \hline
Noisy MNIST & CNN & 1.15$\pm$0.00 & 1.59$\pm$0.01 & \textbf{0.80}$\pm$0.00 & 0.84$\pm$0.00 & 0.85$\pm$0.00 \\ \hline
CIFAR 10 & LR & 62.61$\pm$0.01 & 63.29$\pm$0.01 & 60.99$\pm$0.01 & \textbf{60.73}$\pm$0.01 & 61.13$\pm$0.01 \\ \hline
CIFAR 100 & RN 27 & 34.21$\pm$0.01 & 34.75$\pm$0.01 & 33.82$\pm$0.02 & 33.90$\pm$0.02 & \textbf{33.81}$\pm$0.02 \\ \hline
CIFAR 100 & RN 64 & 31.06$\pm$0.01 & 32.11$\pm$0.02 & \textbf{30.17}$\pm$0.02 & 30.33$\pm$0.02 & 30.51$\pm$0.02 \\ \hline
QT & CNN & 18.59$\pm$0.01&18.52$\pm$0.01&18.68$\pm$0.01&\textbf{18.39}$\pm$0.01&18.48$\pm$0.01   \\ \hline
CoNLL 2003 & CNN & 11.96$\pm$0.02 & 11.85$\pm$0.02 & 12.04$\pm$0.02 & 11.65$\pm$0.02 & \textbf{11.60}$\pm$0.02  \\ \hline
OntoNote 5.0 & CNN & 18.11$\pm$0.02 & 18.03$\pm$0.03 & 18.70$\pm$0.02 & 18.08$\pm$0.02 & \textbf{17.84}$\pm$0.02  \\ \hline
MNIST & FC & 2.91 & \textbf{2.26} & 3.15 & 2.78 & 2.41 \\ \hline
MNIST (distill) & FC & 2.33 & 2.21 & 2.41 & 2.24 & \textbf{2.14} \\ \hline

\end{tabular}
}
\end{table}

\begin{table}[h]
\centering
\caption{Training error rates (Best) of sampling methods (\%)}
\label{tb:sampling_train_results}
\scalebox{0.8}{
\begin{tabular}{cccccccc}
\hline
Datasets & Model & SGD-Uni  & SGD-SD & SGD-ISD & SGD-SE &  SGD-SPV & SGD-STC \\ \hline
MNIST & CNN & 0.01 & \textbf{0.00} & 0.01 & 0.05 & \textbf{0.00} & \textbf{0.00} \\ \hline
Noisy MNIST & CNN & 5.54$\pm$0.09 & \textbf{0.01}$\pm$0.00 & 2.88$\pm$0.09 & 9.08$\pm$0.01 & 7.60$\pm$0.06 & 7.83$\pm$0.04 \\ \hline
CIFAR 10 & LR & 59.88$\pm$0.02 & 60.50$\pm$0.02 & 59.88$\pm$0.02 & 58.49$\pm$0.02 & \textbf{58.26}$\pm$0.02 & 58.42$\pm$0.02 \\ \hline
QT & CNN & \textbf{0.00} & \textbf{0.00} & \textbf{0.00} & 0.04$\pm$0.01 & \textbf{0.00} & \textbf{0.00} \\ \hline

\end{tabular}
}
\end{table}

\begin{table}[h]
\centering
\caption{Training error rates (Best) of sampling methods (\%). For CIFAR 100, the training errors are computed on the randomly cropped and flipped images.}
\label{tb:weighting_train_results}
\scalebox{0.8}{
\begin{tabular}{ccccccc}
\hline
Datasets & Model & SGD-Scan  & SGD-WD & SGD-WE &  SGD-WPV & SGD-WTC \\ \hline
MNIST & CNN & 0.01 & \textbf{0.00} &  0.04 & 0.01 & 0.01 \\ \hline
Noisy MNIST & CNN & 6.21$\pm$0.15 & \textbf{0.29}$\pm$0.02 &  9.01$\pm$0.01 & 7.93$\pm$0.05 & 8.02$\pm$0.04 \\ \hline
CIFAR 10 & LR & 59.87$\pm$0.02 & 60.48$\pm$0.02  & 58.45$\pm$0.02 & \textbf{58.23}$\pm$0.02 & 58.40$\pm$0.02 \\ \hline
CIFAR 100 & RN 27 & 18.72$\pm$0.04 & \textbf{18.44}$\pm$0.04  & 19.43$\pm$0.04 & 18.86$\pm$0.04 & 18.76$\pm$0.04 \\ \hline
CIFAR 100 & RN 64 & 6.06$\pm$0.03 & \textbf{5.42}$\pm$0.04  & 8.15$\pm$0.03 & 8.41$\pm$0.03 & 7.85$\pm$0.02 \\ \hline
QT & CNN & \textbf{0.00} & \textbf{0.00} & 0.04$\pm$0.01 & \textbf{0.00} & \textbf{0.00} \\ \hline
CoNLL 2003 & CNN & 2.55$\pm$0.03 & \textbf{1.64}$\pm$0.02 & 4.00$\pm$0.03 & 2.14$\pm$0.01 & 1.86$\pm$0.02 \\ \hline
OntoNote 5.0 & CNN & 13.90$\pm$0.03 & 13.16$\pm$0.05 & 15.21$\pm$0.03 & 13.29$\pm$0.03 & \textbf{12.61}$\pm$0.03  \\ \hline
MNIST & FC & 1.84$\pm$0.01 & \textbf{0.07}$\pm$0.00 & 2.21$\pm$0.02 & 1.60$\pm$0.01 & 0.79$\pm$0.01 \\ \hline
MNIST (distill) & FC & 0.73$\pm$0.01 & \textbf{0.01}$\pm$0.00 & 0.96$\pm$0.01 & 0.58$\pm$0.01 & 0.13$\pm$0.00 \\ \hline

\end{tabular}
}
\end{table}

\section{Proof sketch of Equation~\eqref{eq:WTC}}

In Equation~\eqref{eq:S_N} and~\eqref{eq:var_approx}, by assuming 
\begin{equation}
p(y_i|\mathbf{x_i},W)  \approx p(y_i|\mathbf{x_i},\mathbf{w})+ g_i(\mathbf{w})^T ( W-\mathbf{w} ),
\end{equation}
and 
\begin{equation}
Pr(W=\mathbf{w}|Y,X)   \approx \mathcal{N}(\mathbf{w}|\mathbf{w_N},S_N) 
\end{equation}
, we know that 
\begin{equation}
Var( p(y_i|\mathbf{x_i},W) ) \approx g_i(\mathbf{w})^T S_N g_i(\mathbf{w}).
\end{equation}

We apply the 
\begin{equation}
v_i=\frac{1}{N_T}\bar{p}_{H_i^{t-1}}(y_i|\mathbf{x_i})(1-\bar{p}_{H_i^{t-1}}(y_i|\mathbf{x_i}))+ \epsilon_T
\end{equation} to the loss function, so 
\begin{equation}
L = - \sum_i v_i \log( p(y_i|\mathbf{x_i},\mathbf{w}) ) - \frac{c}{s_0} ||\mathbf{w}||^2.
\end{equation} Then, $S_N^{-1}=\sum_i v_i p(y_i|\mathbf{x_i})\left(1-p(y_i|\mathbf{x_i})\right) \mathbf{x_i}\mathbf{x_i}^T + \frac{2c}{s_0}I$.

When $p(y_i|\mathbf{x_i},\mathbf{w})\approx \bar{p}_{H_i^{t-1}}(y_i|\mathbf{x_i})$, $1/s_0=0$, and $\epsilon_T=0$, 
\begin{align}
& Tr(g_i(\mathbf{w})^T S_N g_i(\mathbf{w}))
= Tr(g_i(\mathbf{w}) g_i(\mathbf{w})^T S_N) \nonumber \\
& \approx Tr\left( \left( p(y_i|\mathbf{x_i},\mathbf{w})^2\left(1-p(y_i|\mathbf{x_i},\mathbf{w})\right)^2\mathbf{x_i} \mathbf{x_i}^T \right)  \left( \frac{1}{N_T} \sum_j p(y_j|\mathbf{x_j},\mathbf{w})^2\left(1-p(y_j|\mathbf{x_j},\mathbf{w})\right)^2 \mathbf{x_j}\mathbf{x_j}^T \right)^{-1}  \right)
\end{align}

Finally, 
\begin{align}
& \sum_i g_i(\mathbf{w})^T S_N g_i(\mathbf{w})=\sum_i Tr(g_i(\mathbf{w})^T S_N g_i(\mathbf{w})) \nonumber \\
& \approx \sum_i Tr\left( p(y_i|\mathbf{x_i},\mathbf{w})^2\left(1-p(y_i|\mathbf{x_i},\mathbf{w})\right)^2\mathbf{x_i} \mathbf{x_i}^T \left( \frac{1}{N_T} \sum_j p(y_j|\mathbf{x_j},\mathbf{w})^2\left(1-p(y_j|\mathbf{x_j},\mathbf{w})\right)^2 \mathbf{x_j}\mathbf{x_j}^T \right)^{-1} \right) \nonumber \\
& = N_T Tr(I)= N_T \cdot dim(w).
\end{align}

\end{document}